\documentclass[10pt,twocolumn,letterpaper]{article}

\usepackage{cvpr}              % To produce the CAMERA-READY version

\usepackage{algorithm}
\usepackage{algorithmic}

\usepackage{amsmath}
\usepackage{booktabs}
\usepackage{multirow}
\usepackage{threeparttable}
\usepackage{adjustbox}
\usepackage{pifont}
\usepackage[switch]{lineno}

\usepackage{epsfig}
\usepackage{rotating}
\usepackage{bbding}
\usepackage{enumitem}
\usepackage{bm}
\usepackage{colortbl}
\usepackage{graphicx}
\usepackage{subcaption}
\usepackage{tabularx}
\usepackage{booktabs}
\usepackage[accsupp]{axessibility}

\usepackage[dvipsnames]{xcolor}

\definecolor{cvprblue}{rgb}{0.21,0.49,0.74}
\usepackage[pagebackref,breaklinks,colorlinks,citecolor=cvprblue]{hyperref}

\def\paperID{13298} % *** Enter the Paper ID here
\def\confName{CVPR}
\def\confYear{2024}

\title{Leveraging Predicate and Triplet Learning for Scene Graph Generation}

\author{Jiankai Li$^{1,2,3}$ \hspace{10pt} Yunhong Wang$^{1}$ \hspace{10pt} Xiefan Guo$^{1}$ \hspace{10pt} Ruijie Yang$^{1}$ \hspace{10pt} Weixin Li$^{1,2,3}$\thanks{Corresponding author}\\
$^1$ IRIP Lab, School of Computer Science and Engineering, Beihang University, Beijing, China \\
$^2$ State Key Laboratory of Complex \& Critical Software Environment, Beihang University, Beijing, China\\
$^3$ Shanghai Artificial Intelligence Laboratory, Shanghai, China\\
{\tt\small \{lijiankai, yhwang, xfguo, rjyang, weixinli\}@buaa.edu.cn}
}

\begin{document}
\maketitle
\begin{abstract}

Scene Graph Generation (SGG) aims to identify entities and predict the relationship triplets \textit{\textless subject, predicate, object\textgreater } in visual scenes. Given the prevalence of large visual variations of subject-object pairs even in the same predicate, it can be quite challenging to model and refine predicate representations directly across such pairs, which is however a common strategy adopted by most existing SGG methods. We observe that visual variations within the identical triplet are relatively small and certain relation cues are shared in the same type of triplet, which can potentially facilitate the relation learning in SGG.
Moreover, for the long-tail problem widely studied in SGG task, it is also crucial to deal with the limited types and quantity of triplets in tail predicates. 
Accordingly, in this paper, we propose a Dual-granularity Relation Modeling (DRM) network to leverage fine-grained triplet cues besides the coarse-grained predicate ones.
DRM utilizes contexts and semantics of predicate and triplet with Dual-granularity Constraints, generating compact and balanced representations from two perspectives to facilitate relation recognition.
Furthermore, a Dual-granularity Knowledge Transfer (DKT) strategy is introduced to transfer variation from head predicates/triplets to tail ones, aiming to enrich the pattern diversity of tail classes to alleviate the long-tail problem.
Extensive experiments demonstrate the effectiveness of our method, which establishes new state-of-the-art performance on Visual Genome, Open Image, and GQA datasets.
Our code is available at \url{https://github.com/jkli1998/DRM}
\end{abstract}

\vspace{-1em}
\section{Introduction}
\label{sec:intro}

Entities and their associated relationships form the cornerstones of visual contents in images \cite{krishna2017visual}. 
Scene Graph Generation (SGG), a fundamental task in visual scene understanding, is designed to detect these entities and predict their pairwise relationships, encapsulating them into \textit{\textless subject, predicate, object\textgreater } triplets \cite{lu2016visual,dai2017detecting,zhang2017visual}. 
The generated compact graph-structured image representation can be utilized in a range of applications, \eg embodied navigation \cite{du2020learning,singh2023scene}, image retrieval \cite{dhamo2020semantic,guo2020visual}, visual question answering \cite{ben2019block,shi2019explainable}, \etc, so the SGG task has received widespread attention in recent years \cite{lin2020gps,yang2022panoptic,li2023zero,jung2023devil}.

\begin{figure}
\centering {\includegraphics[width=0.4\textwidth]{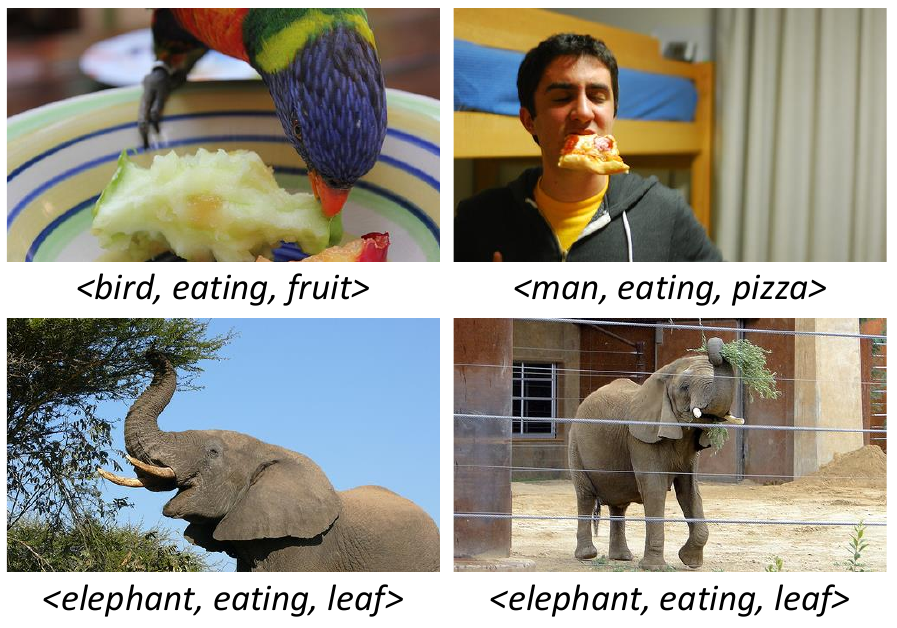}}
\vspace{-0.5em}
\caption{
The illustration of large visual variations within the predicate ``eating''.
Identical predicate can appear differently under distinct subject-object pairs, encompassing a different set of visual cues within each manifestation. 
Identifying discriminative relation cues that are shared across diverse subject-object pairs within the same predicate can be challenging. Yet, they can be easily captured when the scope is narrowed to the identical triplet.
}
\label{fig:intro1}
\vspace{-1em}
\end{figure}

Existing SGG methods are mostly dedicated to generating discriminative predicate representations for the detected entities, based on their appearances, relative positions, contextual cues, \textit{etc.} \cite{yang2018graph,tang2019learning,li2021bipartite,kundu2023ggt,zheng2023prototype}. 
However, as shown in Figure \ref{fig:intro1}, large visual variations due to different subject-object combinations are inherent even in the same predicate, presenting an obstacle for SGG methods to capture robust predicate cues across distinct triplet types.
To alleviate this problem, PE-Net \cite{zheng2023prototype} utilizes textual semantics of predicate categories as the prototype and model  predicate cues by reducing the intra-class variance and inter-class similarity.
Despite the improved prediction accuracy, PE-Net still follows the previous strategy of straightforwardly amalgamating predicate cues that probably contain extensive visual variations of diverse triplets.

Moreover, many recent efforts have been devoted to the long-tail problem in SGG task. Insufficient samples with limited triplet types lead to the reduced diversity observed in tail predicate categories, making it a challenging task to learn and adapt to their distributions. Existing methods boost model's attention towards tail predicate categories through re-sampling \cite{li2021bipartite}, re-weighting\cite{lyu2022fine}, or the utilization of mixture of experts \cite{dong2022stacked,sudhakaran2023vision}. 
However, they mostly fall short of directly tackling the core of the long-tail issue, \ie, the insufficient patterns for tail predicates, leaving space for further improvement.

Reflecting on similarities and differences among predicates, we find that despite the non-negligible or even large variations inherent in the same predicate, the visual diversity of the same triplet is relatively small (\eg the two instances of 
\textit{\textless elephant, eating, leaf\textgreater } in Figure \ref{fig:intro1}). 
Accordingly, considering the more fine-grained triplet cues in addition to the coarse-grained predicate ones can help prevent the model from getting stuck in the refinement process of predicate features with potentially large variations, and promote it to strike a balance between different granularities during the relationship learning process.
For the long-tail problem in SGG task which primarily emerges due to the insufficient tail predicate patterns and corresponding limited types and quantity of triplets, it then becomes a natural choice to enrich tail predicate patterns using head predicates and their triplets.

\begin{figure}
\centering {\includegraphics[width=0.45\textwidth]{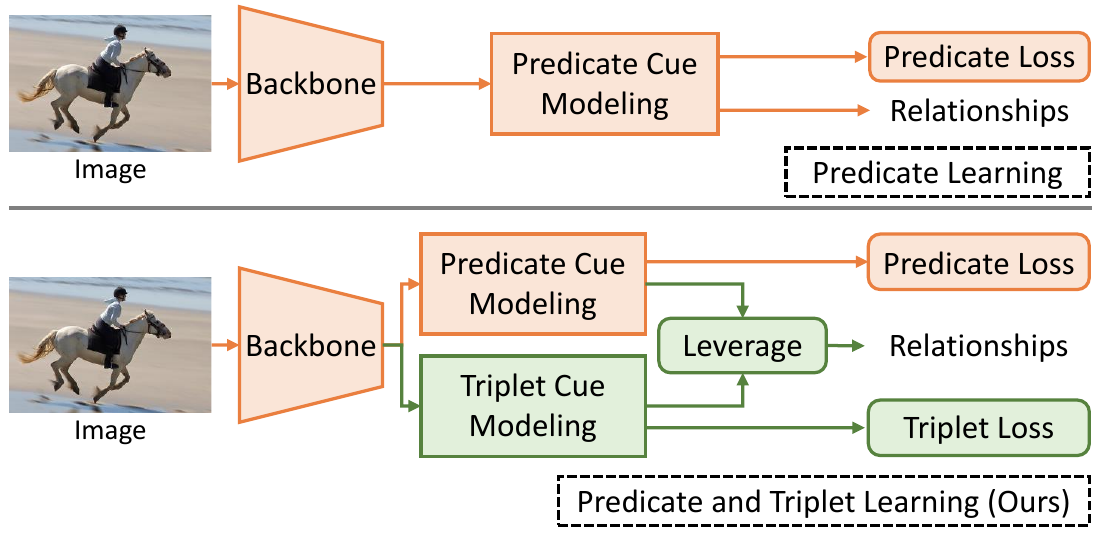}}
\vspace{-0.8em}
\caption{
Comparison of different pipelines for relation recognition. 
Previous methods focus on learning predicate cues shared across various triplets with diverse visual appearance. 
Our method learns and leverages both triplet cues within the same triplet and predicate cues across triplets, to better handle the visual diversity.
}
\label{fig:intro}
\vspace{-1em}
\end{figure}

Based on the aforementioned insights, we propose a Dual-granularity Relation Modeling (DRM) network that models triplet cues to facilitate predicate learning and transfer knowledge from the head classes to tail classes for relation recognition. 
In our DRM network, as shown in Figure \ref{fig:intro}, 1) besides a predicate branch that models coarse-grained predicate cues by leveraging their contextual predicates and subject-object pairs, a triplet branch is also presented to strengthen fine-grained triplet representations via jointly exploring their visual contents and corresponding label semantics. 
We also devise dual-granularity constraints to prevent the degradation of predicate and triplet feature spaces during model training.
Subsequently, 2) the Dual-granularity Knowledge Transfer (DKT) strategy is proposed to transfer the class variance from head classes to tail ones from both the predicate and triplet perspectives for unbiased SGG. 
Distributions of tail predicates/triplets are designed to be calibrated using variance from head ones that are most similar to them. 
New predicate/triplet samples are generated as well based on the calibrated distributions to enrich the pattern diversity of tail predicates/triplets.
Extensive experiments conducted on the widely used Visual Genome \cite{krishna2017visual}, Open Image \cite{krasin2017openimages}, and GQA \cite{hudson2019gqa} datasets for SGG demonstrate the state-of-the-art performance of our method.

Contributions of this paper can be summarized as:
\begin{itemize}
\item We propose to learn scene graphs in a dual-granularity manner, integrating both coarse-grained predicate cues and fine-grained triplet cues.
\item We introduce the DRM network to model predicate and triplet cues, and propose the DKT strategy to propagate knowledge from head predicates/triplets to tail ones to mitigate the long-tail problem in SGG.
\item We comprehensively evaluate the proposed method and demonstrate its superior performance.
\end{itemize}

\begin{figure*}[!tb]
\centering {\includegraphics[width=1.0\textwidth]{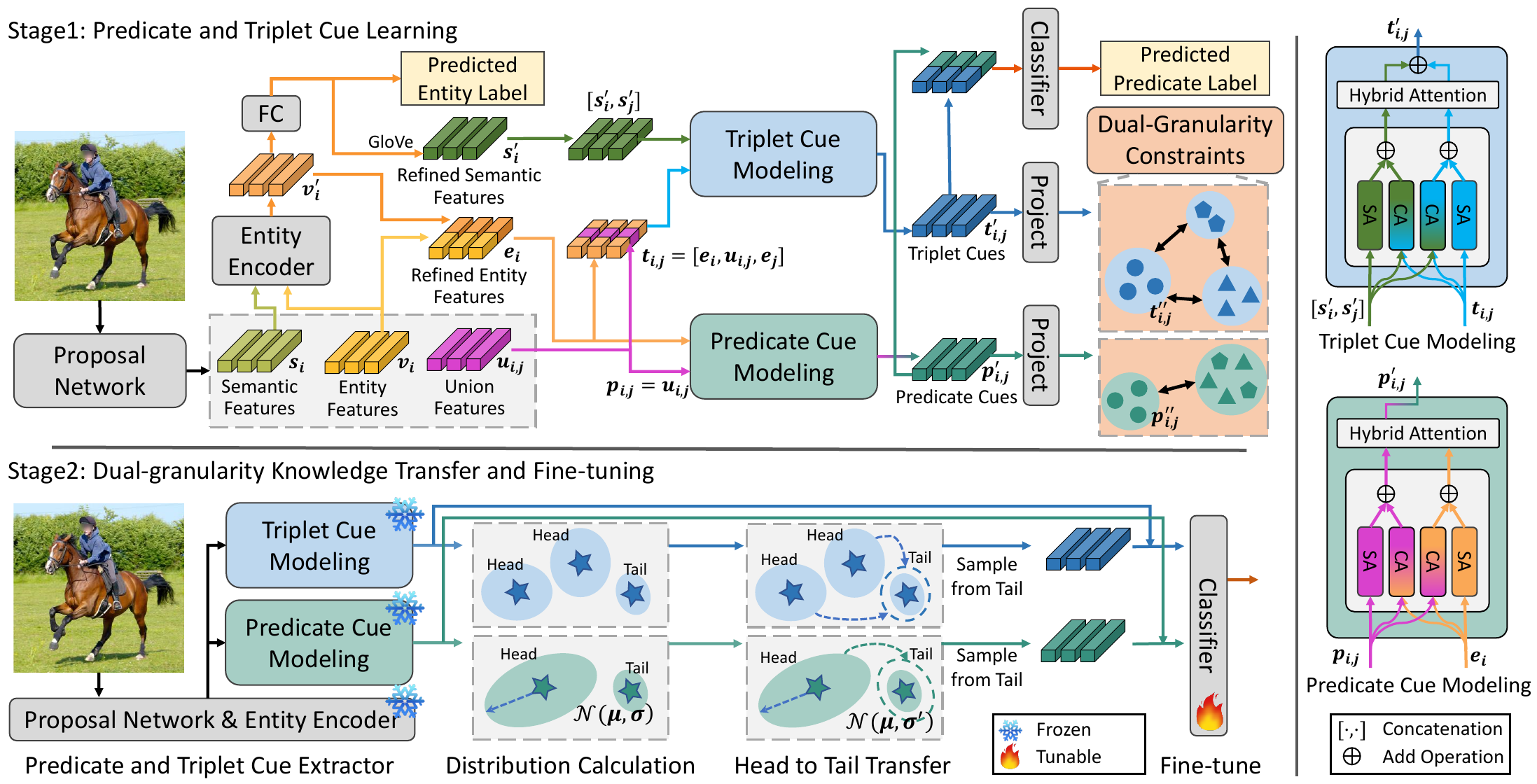}}
\vspace{-1.5em}
\caption{
Illustration of the proposed Dual-granularity Relation Modeling (DRM) network. 
The learning procedure of DRM is composed of two stages. 
In the first stage, we capture the coarse-grained predicate cues shared across different subject-object pairs and learn the fine-grained triplet cues under specific subject-object pairs.
In the second stage, the Dual-granularity Knowledge Transfer (DKT) strategy transfers the variation from head predicates with their associate triplets to the tail. 
Then DRM exploits the real instances along with synthetic samples from the calibrated tail distribution to fine-tune the relation classifier, which alleviates the long-tail problem in SGG.
}
\label{fig:framework}
\vspace{-1em}
\end{figure*}

\section{Related Work}
\label{sec:related work}

\subsection{Scene Graph Generation}

Scene graph is a structured representation of the image content, where its essential constituents are the relationships (or triplets \textit{\textless subject, predicate, object\textgreater }). 
The direct prediction of triplet categories presents a significant challenge given the extensive combinations of subjects, predicates, and objects.
Existing Scene Graph Generation (SGG) methods \cite{woo2018linknet,zhang2019graphical,liu2021fully,li2023zero,zheng2023prototype,jin2023fast} decompose this prediction target into two components: entities and predicates. 
Early SGG approaches \cite{lu2016visual,zhang2017visual,zellers2018neural} explore to integrate multiple modalities like positional information and linguistic features into relationships. 
Later methodologies \cite{suhail2021energy,tian2021mask,lu2021context,jin2023fast,sun2023unbiased} identify the value of visual context in SGG. 
Some of them encode contextual information utilizing techniques \eg message passing \cite{xu2017scene}, LSTM \cite{zellers2018neural,tang2019learning,wang2020sketching}, graph neural networks \cite{yang2018graph,li2021bipartite}, and self-attention modules \cite{dong2022stacked,chen2022reltransformer,li2022sgtr}.
Others \cite{lu2021context,tian2021mask} refine the detected scene graph and optimize the features of refined predicates based on high-confidence predictions.
PE-Net \cite{zheng2023prototype} proposes to utilize text embeddings as the centroid of predicates, aiming to minimize the intra-class variance and the inter-class similarity.
Despite the improved prediction accuracy, it is still hard to extract discriminative relation cues across the various subject-object pairs in the same predicate.
The core of this issue lies in the fact that the identical predicates can manifest differently under distinct subject-object pairs, encompassing a unique array of visual cues within each manifestation. 
The direct aggregation of predicate features tends to overlook the inherent triplet cues present under the various subject-object pairs. 
We propose to explicitly model both coarse-grained predicate cues and fine-grained triplet cues for biased and unbiased SGG.

\subsection{Unbiased Scene Graph Generation}

Real-world data tends to obey a long-tailed distribution, and imbalanced samples in scene graph present the challenge to learn and adapt to the distribution in the tail predicates \cite{tang2019learning,tang2020unbiased,zhang2023deep,yu2023visually}.
Recently, many methods have been proposed to deal with the biased prediction problem in scene graphs, which can be roughly divided into three categories.
The first category of methods use re-balancing strategies that alleviate the long-tail problem by re-sampling images and triplet samples to enhance the performance of tail predicates \cite{li2021bipartite} or by enhancing the loss weights of tail predicates and easily misclassified categories through the pre-defined Predicate Lattice \cite{lyu2022fine}.
The second category of methods utilize noisy label learning to explicitly re-label relational triplets missed by annotators and correct the mislabeled predicates \cite{zhang2022fine,li2022devil}.
The last category exploit the mixture of experts to let different experts separately deal with a sub-part of predicates and merge their outputs \cite{dong2022stacked,sudhakaran2023vision}.
Different from these approaches, we mitigate the long-tail problem in SGG by exploiting knowledge from the head predicates and triplets.
Our method involves transferring the abundant variations from the head predicates and their associate triplets to the tail, thus enriching patterns of the tail predicates.

% However, these methods do not address the fundamental issue of the biased SGG problem, that is, the scarcity of samples in the tail predicates. 
% Additionally, they also neglect the inherent internal structure within predicates.
% TDE \cite{tang2020unbiased} proposes to use counterfactual causality to mitigate contextual bias. 
% EBM \cite{suhail2021energy} suggests the employment of energy model to increase the similarity between generated visual scene graphs and textual ground truth for an unbiased prediction.
% SHA+GCL \cite{dong2022stacked} introduces a divide and conquer strategy. It divides the imbalanced predicates into several groups and then employs a mixture of experts to collaborate on these groups.
% CV-SGG \cite{jin2023fast} improves unbiased prediction by perturbating the sizes and positions of the subject-object pairs to weaken the role of vision.

\section{Method}

An overview of our Dual-granularity Relation Modeling (DRM) network is illustrated in Figure \ref{fig:framework}. 
DRM aims to balance and integrate coarse-grained predicate cues and fine-grained triplet cues for relation recognition. 
To this end, a predicate cue modeling module and a triplet cue modeling module are designed to extract predicate cues shared across diverse subject and object pairs, and triplet cues shared by specific subject-object pairs, respectively. 
Moreover, the Dual-granularity Knowledge Transfer (DKT) strategy is proposed for unbiased SGG. 
This strategy transfers the knowledge of both predicates and triplets from head categories to tail ones, with the objective of enriching the pattern of tail predicates and their associated triplets.

\subsection{The DRM Network Backbone}
The backbone of our DRM network is composed of a proposal network and an entity encoder. Features generated by the backbone are further fed into subsequent predicate and triplet cue modeling modules.

\textbf{Proposal Network.} 
Given an image $\mathcal{I}$, the proposal network generates $N$ entities
along with their corresponding visual features, label predictions, and spatial features from their bounding boxes.
A pre-trained Faster RCNN \cite{ren2015faster} is adopted as the proposal network in this paper.
And following previous works \cite{zellers2018neural,tang2019learning,li2021bipartite,dong2022stacked}, we initialize the entity representations $\{\mathbf{v}_i\}_{i=1}^N$ with their visual and spatial features, encode the union feature $\mathbf{u}_{i,j}$ between the $i$-th entity and the $j$-th entity with their relative spatial representation and the ROI feature of their union box, and obtain the semantic features $\{\mathbf{s}_i\}_{i=1}^N$ of entities using word embeddings of their class labels.

\textbf{Entity Encoder.} The entity encoder is designed to refine features of entities with their contexts for further predictions. 
Inspired by Dong \textit{et al.} \cite{dong2022stacked}, we utilize the Hybrid Attention (HA) to incorporate semantic cues $\{\mathbf{s}_i\}_{i=1}^N$ into entities $\{\mathbf{v}_i\}_{i=1}^N$ while modeling the scene context. 
Each layer of Hybrid Attention is composed of two Self-Attention (SA) units and two Cross-Attention (CA) units, which is built upon the multi-head attention module \cite{vaswani2017attention}.
The Hybrid Attention at the $l$-th layer can be formulated as:
\begin{equation}
    \left \{
    \begin{aligned}
	\mathbf{X}^{(l)} = SA(\mathbf{X}^{(l-1)}) + CA(\mathbf{X}^{(l-1)}, \mathbf{Y}^{(l-1)}),
	\\
	\mathbf{Y}^{(l)} = SA(\mathbf{Y}^{(l-1)}) + CA(\mathbf{Y}^{(l-1)}, \mathbf{X}^{(l-1)}),
	\end{aligned}
	\right.
\end{equation}
where $\mathbf{X}^{(l-1)}$ and $\mathbf{Y}^{(l-1)}$ are inputs of the $l$-th HA layer. We directly fuse outputs of SA and CA units with an addition operation.

Our entity encoder $Enc_{ent}$ is consisted of a stacked 4-layer Hybrid Attention, and we have $\mathbf{X}^{(0)}=\{\mathbf{v}_i\}_{i=1}^N$, $\mathbf{Y}^{(0)}=\{\mathbf{s}_i\}_{i=1}^N$, and:
\begin{equation}
    \{\mathbf{v}^{\prime}_i\}_{i=1}^N = Enc_{ent}(\{\mathbf{v}_i\}_{i=1}^N, \{\mathbf{s}_i\}_{i=1}^N),
\end{equation}
where the refined entity representations $\{\mathbf{v}^{\prime}_i\}_{i=1}^N$ are obtained by summing up of outputs from the last HA layer.

\subsection{Predicate and Triplet Cue Modeling}
Previous approaches tend to categorize predicates in a coarse manner, most of which only focus on predicate cues shared among various subject-object pairs, and thus cannot effectively deal with the potentially large visual variations inherent in the identical predicate.
In contrast to these approaches, our method considers both coarse-grained predicate cues and fine-grained triplet cues, leveraging and striking a balance between the dual-grained cues for accurate predicate categorization. 
Our DRM network involves a Predicate Cue Modeling module and a Triplet Cue Modeling module for extracting features at these two granularities respectively. 
Additionally, we introduce Dual-Granularity Constraints to decrease the intra-class variance and increase inter-class distinguishability, explicitly enforcing the predicate and triplet branches to concentrate on cues from their corresponding granularities and preventing the degradation of dual-granularity space.

\textbf{Predicate Cue Modeling.}
Our predicate cue modeling module $Enc_{prd}$, comprising a 2-layer Hybrid Attention, aims to capture predicate cues across different subject-object pairs. 
In each HA layer of $Enc_{prd}$, the two Self-Attention units are designed to model predicate and entity contextual information. 
And the two Cross-Attention units are designed to capture the dependency between entities and predicates, 
where the predicate $\mathbf{p}_{i,j}$ solely queries the contexts of its corresponding subject $\mathbf{e}_i$ and object $\mathbf{e}_j$, and the entity $\mathbf{e}_i$ only queries predicates related to it.
Our Predicate Cue Modeling process can thus be formulated as:
\begin{equation}
    \{\mathbf{p}^{\prime}_{i,j}\}_{i\neq j}^{M} = Enc_{prd}(\{\mathbf{p}_{i,j}\}_{i\neq j}^{M}, \{\mathbf{e}_i\}_{i=1}^N),
\end{equation}
where the predicate representation $\mathbf{p}_{i,j}$ is initialized with the union feature $\mathbf{u}_{i,j}$, 
the entity representation $\mathbf{e}_i$ is obtained with the concatenation of $\mathbf{v}^{\prime}_i$ and $\mathbf{v}_i$, $M=N\times(N-1)$ is the number of subject-object pairs, and $\mathbf{p}^{\prime}_{i,j}$ is the corresponding output of $\mathbf{p}_{i,j}$ at the last HA layer.

\textbf{Triplet Cue Modeling.}
Our triplet cue modeling module $Enc_{tpt}$ is also constructed using a 2-layer Hybrid Attention. This module is responsible for obtaining fine-grained triplet cues that are shared by specific subject-object pairs.
The two Self-Attention units in each HA layer of $Enc_{tpt}$ are designed to model visual and semantic contextual information of triplets, respectively, and the two Cross-Attention units aim at fusing semantic cues into the visual information of triplets.
We initialize the triplet representation $\mathbf{t}_{i,j}$ with the concatenation of the subject representation $\mathbf{e}_i$, predicate representation $\mathbf{p}_{i,j}$, and object representation $\mathbf{e}_j$.
Our Triplet Cue Modeling process is formulated as:
\begin{equation}
    \{\mathbf{t}^{\prime}_{i,j}\}_{i\neq j}^{M} = Enc_{tpt}(\{\mathbf{t}_{i,j}\}_{i\neq j}^{M}, \{[\mathbf{s}^{\prime}_i, \mathbf{s}^{\prime}_j]\}_{i\neq j}^{M}),
\end{equation}
where $\mathbf{s}^{\prime}_i$ is the word embedding of predicted entity label, 
and $[\cdot,\cdot]$ denotes the concatenate operation. $\mathbf{t}^{\prime}_{i,j}$ is the contextually and semantically aware triplet feature, and is derived from the addition of outputs from the last HA layer.

\textbf{Dual-granularity Constraints.}
Although we explicitly model predicate and triplet cues with $Enc_{prd}$ and $Enc_{tpt}$, they may degrade beyond our desires with a single predicate cross-entropy loss.
To prevent this degradation, we propose the dual-granularity constraints to guide the predicate and triplet cue modeling modules to refine representations in their desired granularities. 
Specifically, we generate two views of an input relation and impose a predicate category-aware supervised contrastive learning loss on predicate representations to capture the coarse-grained predicate cues as:
\begin{equation}
\label{eq:loss_prd}
  \mathcal{L}_{p} =  - \log{\frac{\exp(\langle \mathbf{p}^{\prime\prime}_{i,j}, \mathbf{p}^{\prime\prime}_{pos_p} \rangle / \tau_{p})}{
  \sum_{b \in \mathcal{B}(i,j)} \exp(\langle \mathbf{p}^{\prime\prime}_{i,j}, \mathbf{p}^{\prime\prime}_{b} \rangle / \tau_{p})}},
\end{equation}
where $\mathbf{p}^{\prime\prime}_{i,j}$ is obtained by passing $\mathbf{p}^{\prime}_{i,j}$ through a projection layer comprising two fully connected layers,
$\tau_{p}$ is the temperature, 
$pos_p$ denotes the subscripts of positive samples belonging to the same predicate category as $\mathbf{p}^{\prime\prime}_{i,j}$, $\langle\cdot,\cdot\rangle$ is the cosine similarity function, 
and $\mathcal{B}(i,j)$ denotes the subscript set of samples within the same batch.

Similarly, a triplet category-aware supervised contrastive learning loss is applied on triplet representations, aiming at capturing the fine-grained triplet cues as:
\begin{equation}
\label{eq:loss_tpt}
  \mathcal{L}_{t} =  -  \log{\frac{\exp(\langle \mathbf{t}^{\prime\prime}_{i,j}, \mathbf{t}^{\prime\prime}_{pos_t} \rangle / \tau_{t})}{
  \sum_{b \in \mathcal{B}(i,j)} \exp(\langle \mathbf{t}^{\prime\prime}_{i,j}, \mathbf{t}^{\prime\prime}_{b} \rangle / \tau_{t})}},
\end{equation}
where $\tau_{t}$ is the temperature, 
$pos_t$ denotes subscripts of positive samples from the same triplet category as $\mathbf{t}^{\prime\prime}_{i,j}$, and $\mathbf{t}^{\prime\prime}_{i,j}$ is obtained by passing $\mathbf{t}^{\prime}_{i,j}$ through a projection layer.

\textbf{Scene Graph Prediction.}
For each relationship proposal, our predicate classifier utilizes two fully connected layers to integrate the coarse-grained and fine-grained cues, \ie $\mathbf{p}^{\prime}_{i,j}$ and $\mathbf{t}^{\prime}_{i,j}$, to obtain the final relation label prediction.
For each entity, we also introduce a fully connected layer with softmax function to get its refined label prediction. 

\textbf{Training Loss.}
During the training in the first stage of DRM, the overall loss function $\mathcal{L}$ is defined as:
\begin{equation}
\label{eq:loss_all}
  \mathcal{L} =  \lambda_{e}\mathcal{L}_{e} + \lambda_{r}\mathcal{L}_{r} + \lambda_{p}\mathcal{L}_{p} + \lambda_{t}\mathcal{L}_{t},
\end{equation}
where $\mathcal{L}_{e}$ and $\mathcal{L}_{r}$ are cross-entropy losses of entities and relationships respectively. $\lambda_{e}$, $\lambda_{r}$, $\lambda_{p}$, and $\lambda_{t}$ are pre-defined weight hyper-parameters for corresponding loss terms.

\subsection{Dual-granularity Knowledge Transfer}
The SGG task typically suffers from the long-tail distribution problem.
This problem primarily originates from the tail predicate classes, which possesses limited types and quantities of triplets. 
The head predicate classes contain a relatively larger number of samples and an abundance of triplet types.
Thus to alleviate the long-tail problem, we propose the Dual-granularity Knowledge Transfer (DKT) strategy to transfer knowledge in predicate and triplet feature spaces from head predicate class to the tail one.
DKT generates samples belonging to the tail predicates with their associated predicate and triplet features, thereby enriching and diversifying the patterns of tail predicates.

Specifically, DKT first calculates the distributions of predicate and triplet features.
We assume that the distribution of each category $\Omega_c$ follows a multidimensional Gaussian distribution. 
Formally, it can be expressed as $\{ \Omega_c = \mathcal{N}(\bm{\mu}_c, \bm{\sigma}_c) | c \in \mathcal{C} \}$, where $\bm{\mu}_c$ and $\bm{\sigma}_c$ denote the mean and covariance of $\Omega_c$, and $c$ denotes the predicate or triplet category.
After the first-stage pre-training of DRM network, we freeze the proposal network, entity encoder, and the predicate and triplet cue modeling modules. 
Subsequently, the compact predicate and triplet features, \ie $\mathbf{p}^{\prime}$ and $\mathbf{t}^{\prime}$, are extracted to calculate the predicate and triplet feature distributions, respectively.
The mean, denoted as $\bm{\mu}_c$, is calculated as $\bm{\mu}_c = \frac{1}{N_c} \sum^{N_c}_{k}\mathbf{x}_k^c$.
The covariance, symbolized as $\bm{\sigma}_c$, is calculated as $\bm{\sigma}_c = \frac{1}{N_c - 1} \sum^{N_c}_k (\mathbf{x}^c_k - \bm{\mu}_c)(\mathbf{x}^c_k - \bm{\mu}_c)^T$.
Here, $\mathbf{x}_i^c$ denotes the feature of category $c$, and $N_c$ is the number of $\mathbf{x}_k^c$. The feature $\mathbf{x}$ can be either $\mathbf{p}^{\prime}$ or $\mathbf{t}^{\prime}$.

We then transfer knowledge of feature distributions of head predicate classes to the tail ones.
Specifically, we arrange the predicate classes in descending order based on their sample numbers, choosing half of the predicate classes as head predicates and the remaining ones as tail predicates. 
We further select triplets that appear more than certain times in the head predicates to be the head triplets (we use 64 times as the threshold in this paper), and those in the tail predicates to be the tail triplets.
For each tail category $i \in \mathcal{C}$, we compute the euclidean distance $d_{i,j}$ between its center $\bm{\mu}_i$ and the center $\bm{\mu}_j$ of head category $j \in \mathcal{C}$.
The closer the centers of two categories are to each other in either predicate or triplet space, the more similar they are, which also increases the likelihood of knowledge sharing between them.
Based on $d_{i,j}$, we achieve the knowledge transfer as:
\begin{equation}
\label{eq:transfer_cov}
  \bm{\sigma}^{\prime}_{i} = \frac{N_{i}}{Q_{i}}\bm{\sigma}_{i} + (1-\frac{N_{i}}{Q_{i}})\sum_{j}\alpha_{i,j}\bm{\sigma}_{j},
\end{equation}
where $\alpha_{i,j}$ denotes the softmax normalized form of $d_{i,j}$, 
$Q_{i}$ denotes the desired number of predicate/triplet instances of tail class $i$ and it is identical for every predicates in the tail. 
It suggests that the tail category with fewer samples requires more knowledge from the head for calibration. 

After dual-granularity knowledge transfer, we generate synthetic samples in tail predicate using corresponding predicate and triplet features from calibrated distributions:
\begin{equation}
\label{eq:sample}
  \{ \Tilde{\bm{x}} | \Tilde{\bm{x}} \sim \Omega^{\prime}_c = \mathcal{N}(\bm{\mu}_c, \bm{\sigma}^{\prime}_c) \}.
\end{equation}

Finally, we under-sample head predicates to form a balanced dataset and input the real instances together with synthetic ones into the relation classifier for fine-tuning. Benefiting from the dual-granularity knowledge transfer, the patterns of tail predicates and their associate triplets can be enriched, thus alleviating the long-tail problem.

\begin{table*}[tbp]
    \centering
    \scalebox{0.9}{
            \begin{tabular}{ l | c c  | c   c | c  c}
                \hline
                \multirow{2}{*}{Models} & \multicolumn{2}{c|}{PredCls} & \multicolumn{2}{c|}{SGCls} & \multicolumn{2}{c}{SGDet} \\
                & R@50/100 & mR@50/100 
               & R@50/100 & mR@50/100 
               & R@50/100 & mR@50/100  \\
                \hline
                \hline
                IMP~\cite{xu2017scene}$_{\textit{CVPR'17}}$
                & 61.1 / 63.1 & 11.0 / 11.8 
                & 37.5 / 38.5 & 6.2 / 6.5 
                & 25.9 / 31.2 & 4.2 / 5.3  \\
                VTransE~\cite{zhang2017visual}$_{\textit{CVPR'17}}$
                & 65.7 / 67.6 & 14.7 / 15.8
                & 38.6 / 39.4 & 8.2 / 8.7 
                & 29.7 / 34.3 & 5.0 / 6.1  \\
                MOTIFS~\cite{zellers2018neural}$_{\textit{CVPR'18}}$
                & 66.0 / 67.9 & 14.6 / 15.8 
                & 39.1 / 39.9 & 8.0 / 8.5
                & 32.1 / 36.9 & 5.5 / 6.8  \\
                G-RCNN~\cite{yang2018graph}$_{\textit{ECCV'18}}$
                & 65.4 / 67.2 & 16.4 / 17.2
                & 37.0 / 38.5 & 9.0 / 9.5 
                & 29.7 / 32.8 & 5.8 / 6.6 \\
                VCTREE~\cite{tang2019learning}$_{\textit{CVPR'19}}$
                & 65.5 / 67.4 & 16.7 / 17.9 
                & 40.3 / 41.6 & 7.9 / 8.3 
                & 31.9 / 36.0 & 6.4 / 7.3  \\
                GPS-Net~\cite{lin2020gps}$_{\textit{CVPR'20}}$
                & 65.2 / 67.1 & 15.2 / 16.6 
                & 37.8 / 39.2 & 8.5 / 9.1 
                & 31.3 / 35.9 & 6.7 / 8.6  \\
                RU-Net~\cite{lin2022ru}$_{\textit{CVPR'22}}$
                & 67.7 / 69.6 & \, \; - \, / 24.2 
                & 42.4 / 43.3 & \, \; - \, / 14.6 
                & 32.9 / 37.5 & \, \; - \, / 10.8  \\
                HL-Net~\cite{lin2022hl}$_{\textit{CVPR'22}}$
                & 67.0 / 68.9 & \, \; - \, / 22.8
                & \textcolor{blue}{\textbf{\underline{42.6}}} / \textcolor{blue}{\textbf{\underline{43.5}}} 
                & \, \; - \, / 13.5 
                & \textcolor{blue}{\underline{\textbf{33.7}}} / \textcolor{blue}{\underline{\textbf{38.1}}}
                & \, \; - \, / 9.2 \,  \\
                PE-Net(P)~\cite{zheng2023prototype}$_{\textit{CVPR'23}}$
                & \textcolor{blue}{\textbf{\underline{68.2}}} / \textcolor{blue}{\textbf{\underline{70.1}}} 
                & 23.1 / 25.4 
                & 41.3 / 42.3 & 13.1 / 14.8
                & 32.4 / 36.9 & \, 8.9  / 11.0  \\
                VETO~\cite{sudhakaran2023vision}$_{\textit{ICCV'23}}$
                & 64.2 / 66.3 & 22.8 / 24.7 
                & 35.7 / 36.9 & 11.1 / 11.9 
                & 27.5 / 31.5 & 8.1 / 9.5  \\
                \hline
                TDE$^{\diamond}$~\cite{tang2020unbiased}$_{\textit{CVPR'20}}$
                & 46.2 / 51.4 & 25.5 / 29.1 
                & 27.7 / 29.9 & 13.1 / 14.9 
                & 16.9 / 20.3 & 8.2 / 9.8 \\
                CogTree$^{\diamond}$~\cite{yu2021cogtree}$_{\textit{IJCAI'21}}$
                & 35.6 / 36.8 & 26.4 / 29.0 
                & 21.6 / 22.2 & 14.9 / 16.1 
                & 20.0 / 22.1 & 10.4 / 11.8  \\
                BPL-SA$^{\diamond}$~\cite{guo2021general}$_{\textit{ICCV'21}}$
                & 50.7 / 52.5 & 29.7 / 31.7 
                & 30.1 / 31.0 & 16.5 / 17.5 
                & 23.0 / 26.9 & 13.5 / 15.6 \\
                VisualDS$^{\diamond}$~\cite{yao2021visual}$_{\textit{ICCV'21}}$
                & \, \; - \;  /\, \; - \; \, & 16.1 / 17.5 
                & \, \; - \;  /\, \; - \; \, & 9.3 / 9.9 
                & \, \; - \;  /\, \; - \; \, & 7.0 / 8.3 \\
                NICE$^{\diamond}$~\cite{li2022devil}$_{\textit{CVPR'22}}$
                & 55.1 / 57.2 & 29.9 / 32.3
                & 33.1 / 34.0 & 16.6 / 17.9 
                & 27.8 / 31.8 & 12.2 / 14.4 \\
                PPDL$^{\diamond}$~\cite{li2022ppdl}$_{\textit{CVPR'22}}$
                & 47.2 / 47.6  & 32.2 / 33.3 
                & 28.4 / 29.3 & 17.5 / 18.2
                & 21.2 / 23.9 & 11.4 / 13.5\\
                GCL$^{\diamond}$~\cite{dong2022stacked}$_{\textit{CVPR'22}}$
                & 42.7 / 44.4 & 36.1 / 38.2 
                & 26.1 / 27.1 & 20.8 / 21.8 
                & 18.4 / 22.0 & 16.8 / 19.3  \\
                IETrans$^{\diamond}$~\cite{zhang2022fine}$_{\textit{ECCV'22}}$
                & \, \; - \;  /\, \; - \; \, & 35.8 / 39.1 
                & \, \; - \;  /\, \; - \; \, & 21.5 / 22.8 
                & \, \; - \;  /\, \; - \; \, & 15.5 / 18.0  \\
                INF$^{\diamond}$~\cite{biswas2023probabilistic}$_{\textit{CVPR'23}}$
                & 51.5 / 55.1 & 24.7 / 30.7  
                & 32.2 / 33.8 & 14.5 / 17.4 
                & 23.9 / 27.1 & \, 9.4 / 11.7  \\
                CFA$^{\diamond}$~\cite{li2023compositional}$_{\textit{ICCV'23}}$
                & 54.1 / 56.6 & 35.7 / 38.2
                & 34.9 / 36.1 & 17.0 / 18.4 
                & 27.4 / 31.8 & 13.2 / 15.5 \\
                EICR$^{\diamond}$~\cite{min2023environment}$_{\textit{ICCV'23}}$
                & 55.3 / 57.4 & 34.9 / 37.0
                & 34.5 / 35.4 & 20.8 / 21.8
                & 27.9 / 32.2 & 15.5 / 18.2  \\
                BGNN~\cite{li2021bipartite}$_{\textit{CVPR'21}}$
                & 59.2 / 61.3 & 30.4 / 32.9 
                & 37.4 / 38.5 & 14.3 / 16.5 
                & 31.0 / 35.8 & 10.7 / 12.6  \\
                SHA+GCL~\cite{dong2022stacked}$_{\textit{CVPR'22}}$
                & 35.1 / 37.2 
                & 41.6 / \textcolor{blue}{\textbf{\underline{44.1}}} 
                & 22.8 / 23.9 
                & 23.0 / 24.3
                & 14.9 / 18.2  
                & 17.9 / 20.9 \\
                PE-Net~\cite{zheng2023prototype}$_{\textit{CVPR'23}}$
                & 64.9 / 67.2 & 31.5 / 33.8
                & 39.4 / 40.7 & 17.8 / 18.9
                & 30.7 / 35.2 & 12.4 / 14.5 \\
                SQUAT~\cite{jung2023devil}$_{\textit{ICCV'23}}$
                & 55.7 / 57.9 & 30.9 / 33.4 
                & 33.1 / 34.4 & 17.5 / 18.8 
                & 24.5 / 28.9 & 14.1 / 16.5 \\
                CaCao~\cite{yu2023visually}$_{\textit{ICCV'23}}$
                & \, \; - \;  /\, \; - \; \, & \textcolor{blue}{\textbf{\underline{41.7}}} / 43.7 
                & \, \; - \;  /\, \; - \; \, & \textcolor{blue}{\textbf{\underline{24.0}}} / \textcolor{blue}{\textbf{\underline{25.0}}} 
                & \, \; - \;  /\, \; - \; \, & \textcolor{blue}{\textbf{\underline{18.3}}} / \textcolor{blue}{\textbf{\underline{22.1}}} \\
                \hline
                \textbf{DRM} w/o \textbf{DKT} 
                & \textcolor{red}{\textbf{70.2}} / \textcolor{red}{\textbf{72.1}} & 23.3 / 25.6
                & \textcolor{red}{\textbf{44.3}} / \textcolor{red}{\textbf{45.2}} & 13.5 / 14.6 
                & \textcolor{red}{\textbf{34.0}} / \textcolor{red}{\textbf{38.9}} & 9.0 / 11.2  \\
                \textbf{DRM} 
                & 43.9 / 45.8 & \textcolor{red}{\textbf{47.1}} / \textcolor{red}{\textbf{49.6}}
                & 27.5 / 28.4 & \textcolor{red}{\textbf{27.8}} / \textcolor{red}{\textbf{29.2}}
                & 19.0 / 22.9 & \textcolor{red}{\textbf{20.4}} / \textcolor{red}{\textbf{24.1}} \\
                \hline
            \end{tabular}
    }
    \vspace{-0.5em}
    \caption{
    Comparison results with state-of-the-art SGG methods on the VG150 dataset. 
    ``$\diamond$'' denotes the combination of MOTIFS with a model-agnostic unbiasing strategy. The best and second best results under each setting are respectively marked in \textcolor{red}{\textbf{red}} and \textcolor{blue}{\textbf{\underline{underline blue}}}.
    }
    \label{tab:compare_vg}
    \vspace{-0.5em}
\end{table*}

\begin{table}[tbp]
    \centering
    \scalebox{0.9}{
            \begin{tabular}{ l | c |c c| c}
                \hline
                  \multicolumn{1}{c|}{\multirow{2}{*}{Model}} & \multicolumn{1}{c|}{\multirow{2}{*}{R@50}} 
                  & \multicolumn{2}{c|}{WmAP} & \multicolumn{1}{c}{\multirow{2}{*}{${\rm {score}}_{wtd}$}} \\ 
                  \cline{3-4}
                \multicolumn{1}{c|}{} & \multicolumn{1}{c|}{} & rel & \multicolumn{1}{c|}{phr} & \multicolumn{1}{c}{} \\
                \hline
                \hline
                MOTIFS~\cite{zellers2018neural}$_{\textit{CVPR'18}}$
                & 71.6 & 29.9 & 31.6 & 38.9  \\
                G-RCNN~\cite{yang2018graph}$_{\textit{ECCV'18}}$
                & 74.5 & 33.2 & 34.2 & 41.8 \\
                VCTREE~\cite{tang2019learning}$_{\textit{CVPR'19}}$
                & 74.1 & 34.2 & 33.1 & 40.2 \\
                GPS-Net~\cite{lin2020gps}$_{\textit{CVPR'20}}$
                & 74.8 & 32.9 & 34.0 & 41.7 \\
                BGNN~\cite{li2021bipartite}$_{\textit{CVPR'21}}$
                & 75.0 & 33.5 & 34.2 & 42.1 \\
                RU-Net~\cite{lin2022ru}$_{\textit{CVPR'22}}$
                & \textcolor{red}{\textbf{76.9}} & 35.4 & 34.9 & 43.5 \\
                HL-Net~\cite{lin2022hl}$_{\textit{CVPR'22}}$
                & \textcolor{blue}{\textbf{\underline{76.5}}} & 35.1 & 34.7 & 43.2  \\
                PE-Net~\cite{zheng2023prototype}$_{\textit{CVPR'23}}$
                & \textcolor{blue}{\textbf{\underline{76.5}}} & \textcolor{blue}{\textbf{\underline{36.6}}} 
                & \textcolor{blue}{\textbf{\underline{37.4}}} & \textcolor{blue}{\textbf{\underline{44.9}}}  \\
                SQUAT~\cite{jung2023devil}$_{\textit{ICCV'23}}$
                & 75.8 & 34.9 & 35.9 & 43.5 \\
                \hline
                \textbf{DRM} w/o \textbf{DKT} 
                & 75.9 & \textcolor{red}{\textbf{40.5}}
                & \textcolor{red}{\textbf{41.4}} & \textcolor{red}{\textbf{47.9}} \\
                \hline
            \end{tabular}
    }
    \vspace{-0.5em}
    \caption{
    Comparison results with state-of-the-art SGG methods on Open Image V6. 
    The best and second best results under each metric are respectively marked in \textcolor{red}{\textbf{red}} and \textcolor{blue}{\textbf{\underline{underline blue}}}.
    }
    \label{tab:compare_open_image_v6}
    \vspace{-1.0em}
\end{table}

\section{Experiments}

%\subsection{Dataset and Evaluation Settings}
\subsection{Experimental Settings}

\textbf{Datasets.} 
We evaluate our method on three commonly used SGG datasets, namely Visual Genome \cite{krishna2017visual}, Open Image \cite{krasin2017openimages}, and GQA \cite{hudson2019gqa}. 
For the Visual Genome dataset, we adopt the VG150 split following previous approaches \cite{xu2017scene,zellers2018neural,tang2019learning,lin2022hl,min2023environment}, which contains the most frequent 150 object categories and 50 predicate categories. 
We use 70\% of images for training, 30\% images for testing and 5k images from the training set for validation. 
As for Open Image, we apply the Open Image V6 protocol, which has 301 object categories and 31 predicate categories. 
It contains 126,368, 1,183, and 5,322 images for training, validation, and testing, respectively. 
For the GQA dataset, we follow previous works \cite{dong2022stacked,sudhakaran2023vision} and utilize the GQA200 split, which includes 200 object categories and 100 predicate categories.

\begin{table*}[tbp]
    \centering
     \scalebox{0.9}{
            \begin{tabular}{ l | c  c | c   c | c c}
                \hline
                \multirow{2}{*}{Models} & \multicolumn{2}{c|}{PredCls} & \multicolumn{2}{c|}{SGCls} & \multicolumn{2}{c}{SGDet} \\
                & R@50/100 & mR@50/100 
               & R@50/100 & mR@50/100
               & R@50/100 & mR@50/100\\
                \hline
                \hline
                VTransE~\cite{zhang2017visual}$_{\textit{CVPR'17}}$
                & 55.7 / 57.9 & 14.0 / 15.0 
                & 33.4 / 34.2 & 8.1 / 8.7 
                & 27.2 / 30.7 & 5.8 / 6.6 \\
                MOTIFS~\cite{zellers2018neural}$_{\textit{CVPR'18}}$
                & \textcolor{blue}{\textbf{\underline{65.3}}} / \textcolor{blue}{\textbf{\underline{66.8}}} & 16.4 / 17.1
                & \textcolor{blue}{\textbf{\underline{34.2}}} / \textcolor{blue}{\textbf{\underline{34.9}}} & 8.2 / 8.6 
                & \textcolor{blue}{\textbf{\underline{28.9}}} / \textcolor{blue}{\textbf{\underline{33.1}}} & 6.4 / 7.7  \\
                VCTREE~\cite{tang2019learning}$_{\textit{CVPR'19}}$
                & 63.8 / 65.7 & 16.6 / 17.4 
                & 34.1  / 34.8 & 7.9 / 8.3 
                & 28.3 / 31.9 & 6.5 / 7.4  \\
                SHA~\cite{dong2022stacked}$_{\textit{CVPR'22}}$
                & 63.3 / 65.2 & 19.5 / 21.1
                & 32.7 / 33.6 & 8.5 / 9.0 
                & 25.5 / 29.1 & 6.6 / 7.8  \\
                VETO~\cite{sudhakaran2023vision}$_{\textit{ICCV'23}}$
                & 64.5 / 66.0
                & 21.2 / 22.1 
                & 30.4 / 31.5 & 8.6 / 9.1  
                & 26.1 / 29.0 & 7.0 / 8.1  \\
                \hline
                VTransE+GCL~\cite{dong2022stacked}$_{\textit{CVPR'22}}$
                & 35.5 / 37.4 & 30.4 / 32.3 
                & 22.9 / 23.6 & 16.6 / 17.4 
                & 15.3 / 18.0 & 14.7 / 16.4 \\
                MOTIFS+GCL~\cite{dong2022stacked}$_{\textit{CVPR'22}}$
                & 44.5 / 46.2 & 36.7 / 38.1 
                & 23.2 / 24.0 & 17.3 / 18.1 
                & 18.5 / 21.8 & 16.8 / 18.8 \\
                VCTREE+GCL~\cite{dong2022stacked}$_{\textit{CVPR'22}}$
                & 44.8 / 46.6 & 35.4 / 36.7 
                & 23.7 / 24.5 & 17.3 / 18.0 
                & 17.6 / 20.7 & 15.6 / 17.8 \\
                SHA+GCL~\cite{dong2022stacked}$_{\textit{CVPR'22}}$
                & 42.7 / 44.5
                & \textcolor{blue}{\textbf{\underline{41.0}}}  / \textcolor{blue}{\textbf{\underline{42.7}}}
                & 21.4 / 22.2 
                & \textcolor{red}{\textbf{20.6}}  / \textcolor{red}{\textbf{21.3}}
                & 14.8 / 17.9 
                & \textcolor{blue}{\textbf{\underline{17.8}}}  / \textcolor{blue}{\textbf{\underline{20.1}}} \\
                \hline
                \textbf{DRM} w/o \textbf{DKT} 
                & \textcolor{red}{\textbf{66.9}} / \textcolor{red}{\textbf{68.4}} & 18.1 / 19.0
                & \textcolor{red}{\textbf{36.4}} / \textcolor{red}{\textbf{37.2}} & 7.1 / 7.4 
                & \textcolor{red}{\textbf{30.6}} / \textcolor{red}{\textbf{34.6}} & 6.9 / 8.4  \\
                \textbf{DRM}
                & 43.2 / 44.4 & \textcolor{red}{\textbf{41.9}} / \textcolor{red}{\textbf{43.5}}
                & 23.3 / 23.9 & \textcolor{blue}{\textbf{\underline{19.9}}} / \textcolor{blue}{\textbf{\underline{20.7}}}
                & 18.6 / 21.7 & \textcolor{red}{\textbf{18.9}} / \textcolor{red}{\textbf{21.0}}\\
                \hline
            \end{tabular}
    }
    \vspace{-0.5em}
    \caption{
    Comparison results with state-of-the-art SGG methods on the GQA200 dataset.
   The best and second best results under each setting are respectively marked in \textcolor{red}{\textbf{red}} and \textcolor{blue}{\textbf{\underline{underline blue}}}.
    }
    \label{tab:compare_gqa}
    \vspace{-1.0em}
\end{table*}

\textbf{Tasks.} 
We adopt three SGG tasks for evaluation: 
1) Predicate Classification (Predcls) infers the predicates of entity pairs with ground-truth bounding boxes and categories.
2) Scene Graph Classification (SGCls) aims to predict the triplet categories with ground-truth bounding boxes.
3) Scene Graph Detection (SGDet) detects bounding boxes of entity pairs and infers their predicate categories.

\textbf{Evaluation Metrics.} 
We use Recall@K (R@K) and mean Recall@K (mR@K) as evaluation metrics on VG150 and GQA200 datasets, following recent works \cite{dong2022stacked,zheng2023prototype,sudhakaran2023vision}. 
R@K tends to prioritize frequent predicates, while mR@K exhibits a preference for less frequent predicates. 
Results on Open Image dataset are evaluated using Recall@50 (R@50), weighted mean AP of relations ($\text{wmAP}_{rel}$), weighted mean AP of phrase ($\text{wmAP}_{phr}$), and a weighted score of them $score_{wtd}=0.2\times R@50 + 0.4 \times \text{wmAP}_{rel} + 0.4 \times \text{wmAP}_{phr}$, following previous works \cite{zhang2019graphical,li2021bipartite,lin2022hl}. 

\textbf{Implementation Details.} 
Following previous works \cite{dong2022stacked,zheng2023prototype,sudhakaran2023vision}, we adopt the pre-trained Faster RCNN with ResNeXt-101-RPN in the proposal network to detect entities in the image. 
GloVe \cite{pennington2014glove} is applied to embed the semantic features. We set the loss weight parameters $\lambda_r$, $\lambda_e$, $\lambda_t$, and $\lambda_p$ as 3, 0.5, 0.1, and 0.1, respectively.
Temperatures $\tau_p$ and $\tau_t$ are set as 0.2, and 0.1 considering that the predicate feature space exhibits greater variety than its associated triplet one.
We optimize our method via SGD, using an initial leaning late of $10^{-4}$ with a batch size of 16.

\begin{table}[tbp]
    \centering
    \setlength{\tabcolsep}{1mm}
         \scalebox{0.85}{
            \begin{tabular}{c c c c | c c | c c }
                \hline
                \multicolumn{4}{c|}{Module} & \multicolumn{2}{c|}{PredCls} & \multicolumn{2}{c}{SGCls}  \\
                P & T & A & C
                & R@50/100 & mR@50/100 & R@50/100 & mR@50/100 \\
                \hline
                \hline
                ~  & ~ & ~ & ~
                & 56.5/60.4  &  15.5/17.4 & 37.5/39.3 & 10.1/11.3 \\
                \Checkmark  & ~ & ~ & ~
                & 67.6/69.5 &  18.1/19.9 & 41.8/42.7 & 10.8/11.8 \\
                ~  & \Checkmark & ~ & ~
                & 67.6/69.8 & 20.5/22.5 & 42.3/43.2 & 12.6/13.6 \\
                \Checkmark  & \Checkmark & ~ & ~
                & 69.3/71.3 &  20.8/22.8 & 42.8/43.8 & 12.8/13.9 \\
                \Checkmark  & \Checkmark & \Checkmark & ~
                & 69.8/71.6 & 20.4/22.3 & 43.4/44.4 & 12.8/13.8 \\
                \Checkmark  & ~ & \Checkmark & \Checkmark
                & 69.8/71.6 & 22.0/24.0 & 43.4/44.3 & 12.3/13.3 \\
                ~ & \Checkmark & \Checkmark & \Checkmark
                & 69.7/71.5 & 21.9/24.3 & 43.8/44.8 & 12.7/13.9 \\
                \hline
                \Checkmark  & \Checkmark & \Checkmark & \Checkmark
                & \textbf{70.2}/\textbf{72.1} & \textbf{23.3}/\textbf{25.6}
                & \textbf{44.3}/\textbf{45.2} & \textbf{13.5}/\textbf{14.6}\\
                \hline
            \end{tabular}
        }
    \vspace{-0.5em}
    \caption{
    Ablation studies on predicate and triplet cues learning. ``P'', ``T'', ``A'', and ``C'' denote the predicate cue modeling module, 
    triplet cue modeling module, augmentation in dual-granularity constraints, and the dual-granularity constraints loss, respectively.
    }
    \label{tab:ablation_stage1}
\vspace{-0.5em}
\end{table}

\begin{table}[tbp]
    \centering
    \setlength{\tabcolsep}{1.2mm}
         \scalebox{0.9}{
            \begin{tabular}{ c | c c | c c }
                \hline
                \multirow{2}{*}{Module} & \multicolumn{2}{c|}{PredCls} & \multicolumn{2}{c}{SGCls}  \\
                ~
                & R@50/100 & mR@50/100 & R@50/100 & mR@50/100 \\
                \hline
                \hline
                None
                & \textbf{70.2}/\textbf{72.1}  &  23.3/25.6 & \textbf{44.3}/\textbf{45.2} & 13.5/14.6 \\
                DKT-P
                & 42.4/44.2 &  45.0/47.3 & 26.8/27.8 & 26.9/28.1 \\
                DKT-T
                & 40.4/42.2 & 46.1/48.7 & 24.3/25.3 & 26.4/28.0 \\
                \hline
                DKT
                & 43.9/45.8 & \textbf{47.1}/\textbf{49.6}
                & 27.5/28.4 & \textbf{27.8}/\textbf{29.2}\\
                \hline
            \end{tabular}
        }
        \vspace{-0.5em}
    \caption{
    Ablation studies on dual-granularity knowledge transfer. ``DKT-P'' and ``DKT-T'' denote the predicate knowledge transfer and triplet knowledge transfer.}
    \label{tab:ablation_stage2}
    \vspace{-1.5em}
\end{table}

\subsection{Comparison with State-of-the-art Methods}
To evaluate the performance our model, we compare it with several state-of-the-art SGG approaches on Visual Genome, Open Image, and GQA datasets. 
The comparison methods include IMP \cite{xu2017scene}, MOTIFS \cite{zellers2018neural}, VCTREE \cite{tang2019learning}, RU-Net \cite{lin2022ru}, HL-Net \cite{lin2022hl}, PE-Net \cite{zheng2023prototype}, and VETO \cite{sudhakaran2023vision}, which focus on the prediction of every relationship in an image. 
We also compare with methods for unbiased scene graph generation, including TDE \cite{tang2020unbiased}, CogTree \cite{yu2021cogtree}, NICE \cite{li2022devil}, INF \cite{biswas2023probabilistic}, CFA \cite{li2023compositional}, EICR \cite{min2023environment}, BGNN \cite{li2021bipartite}, SHA+GCL \cite{dong2022stacked}, SQUAT \cite{jung2023devil}, and CaCao\cite{yu2023visually}.
In addition, we compare our method with VETO+MEET \cite{sudhakaran2023vision} in the setting without graph constraint in the Supplementary Material.

\begin{figure*}[!tb]
\centering
\begin{subfigure}{.195\textwidth}
  \centering
  \includegraphics[width=\linewidth]{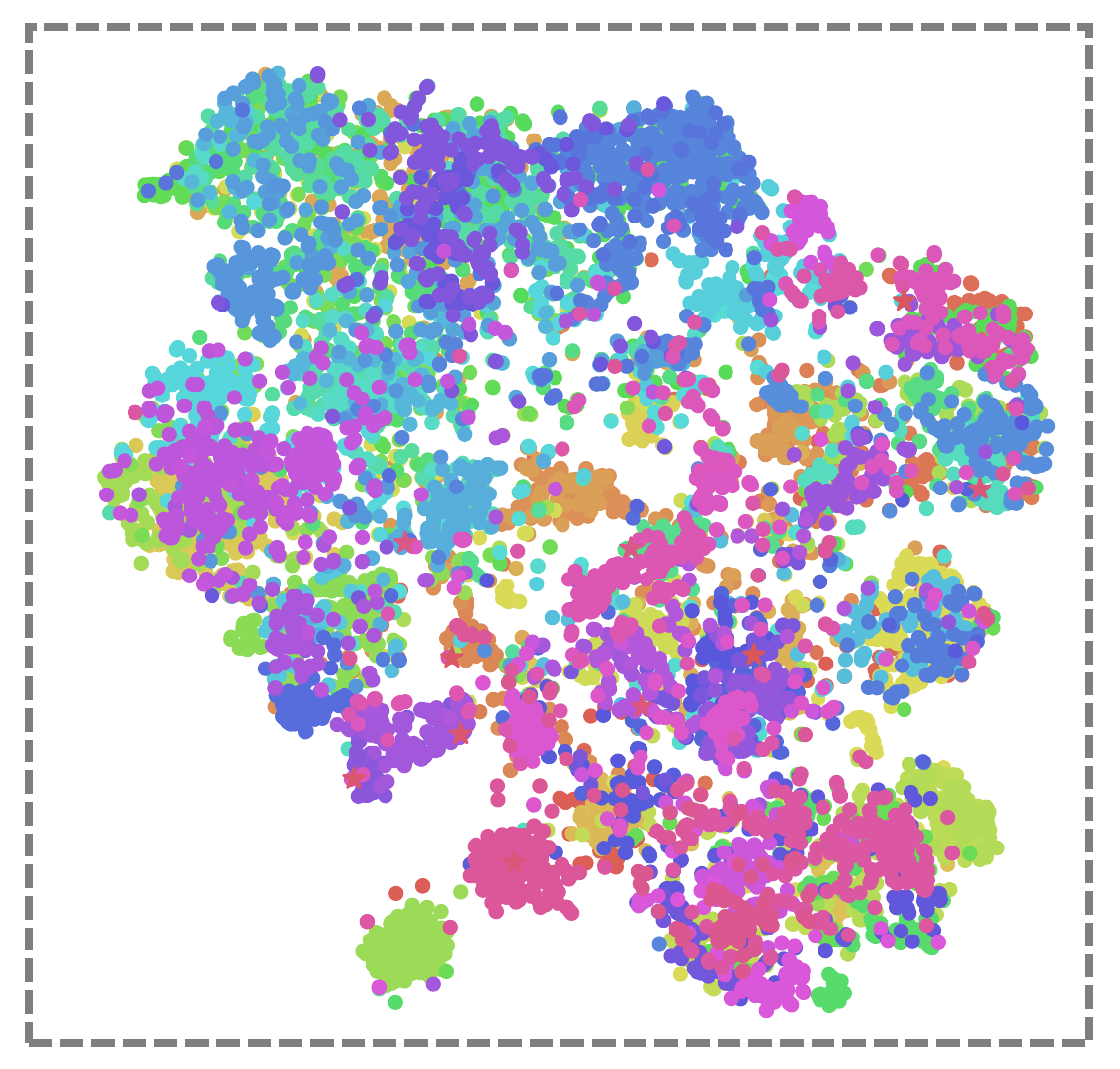}
  \caption{MOTIFS, Triplet}
  \label{fig:tsne1}
\end{subfigure}
\begin{subfigure}{.195\textwidth}
  \centering
  \includegraphics[width=\linewidth]{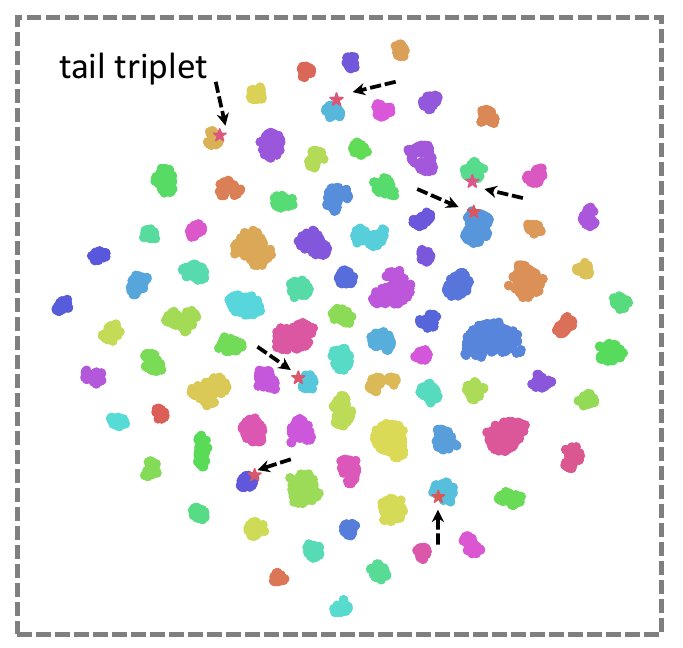}
  \caption{DRM w/o DKT, Triplet}
  \label{fig:tsne2}
\end{subfigure}
\begin{subfigure}{.195\textwidth}
  \centering
  \includegraphics[width=\linewidth]{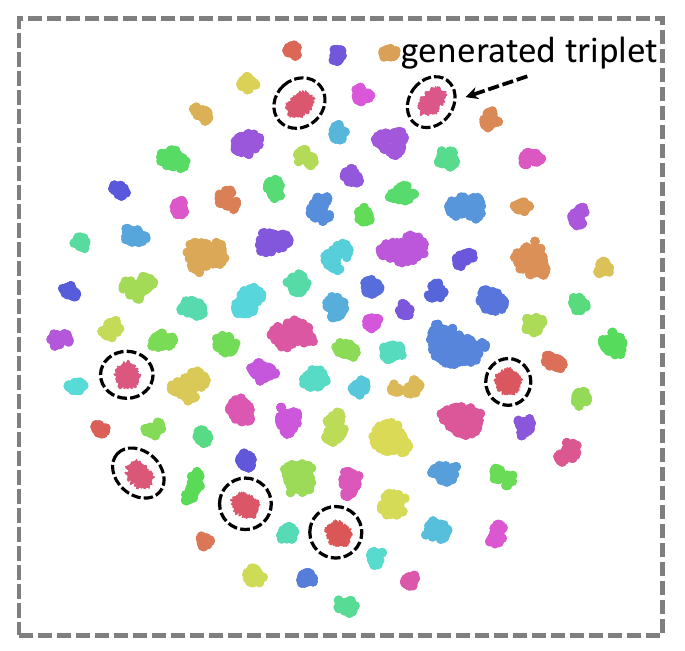}
  \caption{DRM, Triplet}
  \label{fig:tsne3}
\end{subfigure}
\begin{subfigure}{.195\textwidth}
  \centering
  \includegraphics[width=\linewidth]{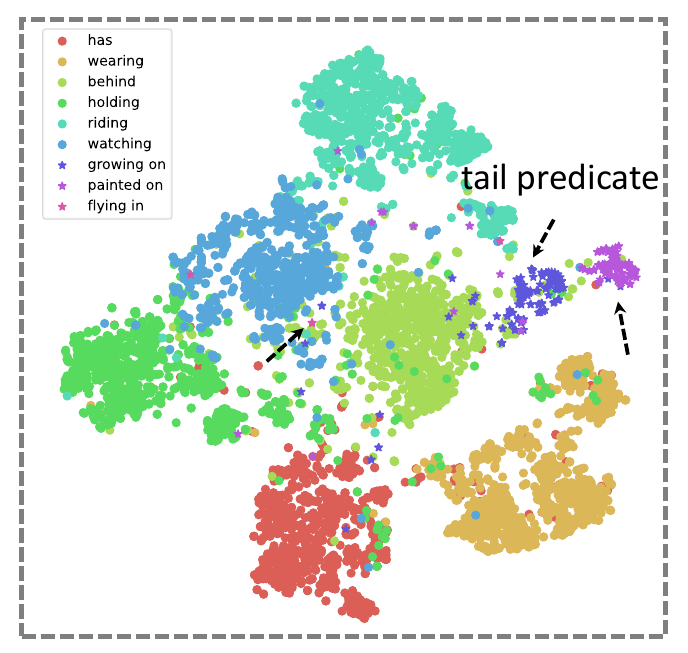}
  \caption{DRM w/o DKT, Predicate}
  \label{fig:tsne4}
\end{subfigure}
\begin{subfigure}{.195\textwidth}
  \centering
  \includegraphics[width=\linewidth]{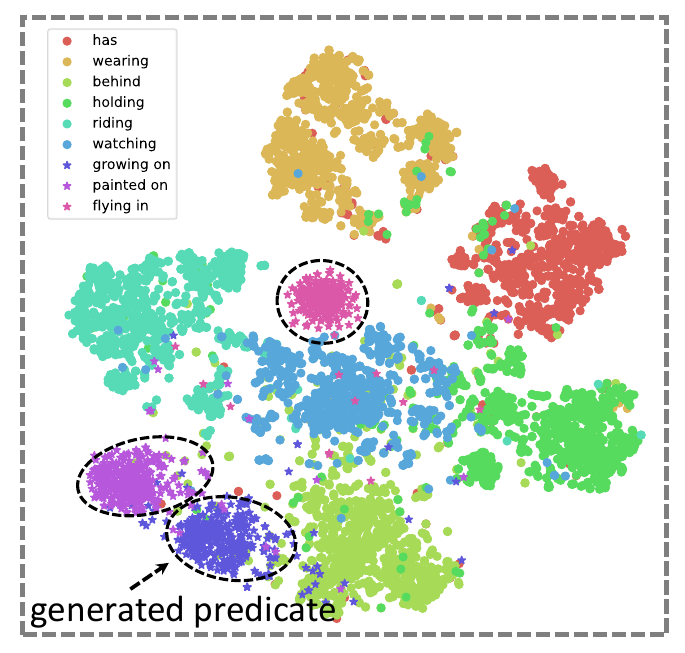}
  \caption{DRM, Predicate}
  \label{fig:tsne5}
\end{subfigure}
\vspace{-0.5em}
\caption{
The comparison of t-SNE visualization results on predicate and triplet feature distributions within the VG dataset.
``MOTIFS, Triplet'' and ``DRM w/o DKT, Triplet'' visualize the same set of samples, where each unique color represents a different type of triplet.
}
\label{fig:tsne}
\vspace{-1.0em}
\end{figure*}

\textbf{Visual Genome.} 
Table \ref{tab:compare_vg} shows the comparison results of different approaches on VG150.
From these results, we have the following observations: 
1) our proposed method significantly outperforms all baselines on all three tasks.
More specifically, our DRM w/o DKT outperforms the recent PE-Net \cite{zheng2023prototype} by 2.0\%, 2.9\%, and 2.0\% at R@100 on PredCls, SGCls, and SGDet, respectively.
% Additionally, it surpasses VCTREE by 4.7\%, 4.6\%, and 2.9\% at R@100 on PredCls, SGCls, and SGDet, respectively. 
Unlike approaches such as MOTIFS \cite{zellers2018neural} and VCTREE \cite{tang2019learning}, which only utilize a predicate classifier to predict predicates and overlook the triplet cues, 
our method leverages coarse-grained predicate cues and fine-grained triplet cues for relation recognition and thus achieves superior results. 
2) Our DRM method also has considerably better performance compared to the baseline unbiased SGG methods. 
Notably, based on the proposed DKT strategy in DRM, our method outperforms the recent multi-expert method SHA+GCL \cite{dong2022stacked} by 5.5\%, 4.9\%, and 3.2\% at mR@100 on three tasks. 
This demonstrates that transferring knowledge from head predicates to tail predicates at dual granularities and enriching the patterns in tail predicates can effectively mitigate the long-tail problem.

\textbf{Open Image.} 
Compared to the VG dataset, Open Image provides a relatively complete labeling of relationships in the images.
Consequently, the model's capability to generate scene graphs can be evaluated at a fine-grained level utilizing the AP metric. 
To evaluate the generalizability of our method across various datasets, we conduct experiments on Open Image V6. 
Since the metrics in Open Image V6 tend to prioritize frequent predicates, we just compare our DRM w/o DKT with state-of-the-art approaches. The comparison results shown in Table \ref{tab:compare_open_image_v6} indicate that our DRM w/o DKT significantly outperforms the recent approaches, \ie, SQUAT \cite{jung2023devil} and PE-Net \cite{zheng2023prototype}. 
% To the best of our knowledge, our work is the first to breakthrough the 40\% precision in both $\text{wmAP}_{rel}$ and $\text{wmAP}_{phr}$ on Open Image V6. 

\textbf{GQA.} 
Compared to Visual Genome and Open Image datasets, GQA200 contains a broader range of predicates.
So we further confirm the generality of our model on the GQA200 dataset. 
As shown in Table \ref{tab:compare_gqa}, our method significantly outperforms the recent VETO \cite{sudhakaran2023vision} at R@100 on three tasks. 
When equipped with DKT, our DRM outperforms SHA+GCL \cite{dong2022stacked} by an average of 0.4\% at mR@100 across three tasks. 
These results demonstrate the consistent effectiveness of our method in handling relation recognition under different data distributions.

\vspace{-0.25em}

\subsection{Ablation Study}
To verify the contributions of different components in our DRM network, we conduct the following ablation studies.

\textbf{Predicate and Triplet Cue Learning.}
We first perform an ablation study on the leverage of predicate and triplet cues.
As shown in Table \ref{tab:ablation_stage1}, we incrementally incorporate one component into the baseline to verify their effectiveness.
% The results exhibit that our modeling of predicate and triplet cues notably enhances performance compared to the direct prediction of predicates.
Compared with the application of predicate or triplet cue modeling in isolation, leveraging both predicate and triplet together improves the performance.
Our dual-granularity constraints include a constraint loss component and a two-view augmentation component to generate positive pairs in each training batch.
The constraints compel predicate and triplet cue modeling to concentrate on corresponding granularities.
We observe from the results that both the loss component and augmentation component contribute to the performance improvements.

\textbf{Dual-Granularity Knowledge Transfer.}
As shown in Table \ref{tab:ablation_stage2}, we conduct an ablation study for the DKT.
We observe an obvious performance gain on mR@K when transferring knowledge at either predicate or triplet granularity.
The mR@K is further improved when predicate and triplet granularities are simultaneously employed.
In comparison to the scenario without the application of DKT, we observe a performance decrease in R@K when using it.
% This is attributed to the capability of our model to reasonably classify ambiguous head predicates, \eg ``on'', into more specific tail predicates, \eg ``sitting on'' and ``walking on''.
% Consequently, decreases at the R@K for these head predicates are inevitable \cite{li2022devil,zheng2023prototype}.
This is attributed to the capability of our model to reasonably classify ambiguous head predicates, \eg ``on'', into more specific tail predicates, \eg ``sitting on''.
Consequently, decreases at the R@K for these head predicates are inevitable \cite{li2022devil,zheng2023prototype}.

\subsection{Visualization Analysis}
To illustrate the ability our method to learn triplet cues and impact of the DKT strategy, we visualize the predicate and triplet feature distributions using t-SNE. 
The visualization results are shown in Figure \ref{fig:tsne}. 
Comparing Figure \ref{fig:tsne1} with Figure \ref{fig:tsne2}, we observe that our DRM generates compact and distinguishable triplet representations, while MOTIFS \cite{zellers2018neural} appears to overlook the triplet cues, leading to a challenge in distinguishing various triplet types.
By comparing Figure \ref{fig:tsne2} with Figure \ref{fig:tsne3}, it can be observed that DKT can generate synthetic samples with diverse distributions for the tail triplets. 
Figures \ref{fig:tsne4} and \ref{fig:tsne5} demonstrate that DKT also transfers the knowledge of head predicates to tail predicates, increasing the tail predicate patterns.
The visualization of predicate and triplet feature distributions demonstrates the interpretability of our method in leveraging predicate and triplet cues and transferring the dual-granularity knowledge.

\section{Conclusion}

In this paper, we propose a Dual-granularity Relation Modeling (DRM) network to address two issues in SGG, \ie the diverse visual appearance within the same predicate and the lack of patterns in tail predicates.
Our DRM network captures the coarse-grained predicate cues shared across different subject-object pairs and fine-grained triplet cues under specific subject-object pairs for relationship recognition.
The Dual-granularity Knowledge Transfer (DKT) is further proposed to transfer the variation from head predicates to the tail to enrich the tail predicate patterns.
Quantitative and qualitative experiments demonstrate that our method establishes new state-of-the art performances on Visual Genome, Open Image and GQA datasets.

\section*{Acknowledgement}

This work was supported in part by the National Key R\&D Program of China under Grant 2022ZD0161901, the National Natural Science Foundation of China under Grants 62276018 and U20B2069, the Beijing Nova Program under Grant 20230484297, the Fundamental Research Funds for the Central Universities, and Research Program of State Key Laboratory of Complex \& Critical Software Environment.

{
    \small
    \bibliographystyle{ieeenat_fullname}
    \bibliography{main}
}

% WARNING: do not forget to delete the supplementary pages from your submission 
\clearpage
\setcounter{page}{1}
\maketitlesupplementary

%\section{Abstract}
In the supplementary material, we provide the following contents for the proposed Dual-granularity Relation Modeling (DRM) network which leverages predicate and triplet learning for Scene Graph Generation (SGG): (1) more implementation details of our method;
(2) more comparison results, including comparisons on the M@K and F@K metrics and comparisons with VETO+MEET \cite{sudhakaran2023vision};
(3) more ablation studies, including the ablation on Dual-granularity Knowledge Transfer (DKT) strategy and dual-granularity learning;
(4) hyper-parameter analysis;
and (5) qualitative visualization. We will make the code publicly available upon acceptance of this paper.

\section{Additional Implementation Details}
We implement DRM using Pytorch \cite{paszke2019pytorch} and the official code-base Scene-Graph-Benchmark.pytorch\footnote{https://github.com/KaihuaTang/Scene-Graph-Benchmark.pytorch} with a NVIDIA A800 GPU. 
In the initialization of entity representations $\{\mathbf{v}_i\}_{i=1}^N$ and union features $\{\mathbf{s}_i\}_{i=1}^N$, we adopt the same strategy as VCTREE \cite{tang2019learning} and PE-Net \cite{zheng2023prototype}, which involves a fusion of their visual and spatial features.
We follow Xu \etal \cite{xu2021end} to augment input images for model training.
The expect number $Q_i$ of each tail predicate in Equation \textcolor{red}{8} is equal to the count of the head predicates with the smallest number.

\section{Additional Comparison Results}

\subsection{Trade-off Results between R@K and mR@K}
Due to the imbalanced data distribution of Visual Genome \cite{krishna2017visual}, Open Image \cite{krasin2017openimages}, and GQA datasets \cite{hudson2019gqa}, there is a trade-off between Recall R@K and mean Recall mR@K metrics. 
To measure the trade-offs of the scene graph generation methods, Zheng \etal \cite{zheng2023prototype} introduce the Mean@K (M@K), which averages the R@K and mR@K, while Zhang \etal \cite{zhang2022fine} propose the F@K, the harmonic mean of R@K and mR@K.
Note that these two metrics only measure the trade-off between R and mR, and it is feasible for diverse methods, even with significant differences in R and mR, to still arrive at the same trade-off results. 
Evaluating either R@K or mR@K better aligns with the practical need to predict the highest number of relationships or to forecast relationships as uniformly as possible.

\noindent\textbf{Visual Genome.} 
Table \ref{tab:compare_vg_mf} shows the performance of different methods in terms of M@50/100 and F@50/100 on VG150. 
Our DRM outperforms state-of-the-art methods at F@K measurements and DRM w/o DTK also achieves state-of-the-art performance at M@K. 
It indicates that although the recall of DRM degrades, the trade-off between the recall and the mean recall is the best in the state-of-the-art methods.

\noindent\textbf{GQA.} 
To evaluate the generalizability of our method across various datasets, we present the performance of M@50/100 and F@50/100 on GQA200.
As shown in Table \ref{tab:compare_gqa_mf}, DRM outperforms all of the state-of-the-art methods at both M@50/100 and F@50/100 metrics.
These results demonstrate that our method remains effective in dealing with relation recognition, regardless of the variations of data distributions.

\subsection{Comparison with VETO+MEET}
The MEET \cite{sudhakaran2023vision} method assigns multiple relationships to each subject-object pair during inference.
This is in accordance with the testing protocol termed ``without graph constraint''.
The setting of ``without graph constraint'', as proposed by Zellers \cite{zellers2018neural}, permits the output scene graph to have multiple edges between the subject and object. 
Better performance is typically achieved without the graph constraint since the model is allowed to make multiple guesses for challenging relations.
In the following, ``ng-’’ denote the No Graph Constraint variant of the metric.

As shown in Table \ref{tab:compare_veto_un}, we compare our DRM with VETO+MEET under the setting of ``without graph constraint'' on VG150 and GQA200 datasets.
We have the following observations:
1) Compared to the performance with graph constraint, our method consistently exhibits promotion without graph constraint.
2) Our proposed DRM w/o DKT has considerably better performance compared to VETO+MEET. More specifically, our DRM w/o DKT outperforms VETO+MEET by an average of 11.9\% and 4.9\% at ng-R@100 and ng-mR@100, respectively.
3) Based on the proposed DKT strategy, DRM significantly outperforms VETO+MEET by 21.1\%, 16.2\%, 16.7\% at ng-mR@100 on three tasks of VG150 datasets. It also surpasses VETO+MEET by 25.2\%, 14.4\%, 14.8\% at ng-mR@100 on three tasks of VG150 datasets.

We also present the comparison results at ng-M@50/100 and ng-F@50/100 to demonstrate the trade-off performance under the setting of ``without graph constraint’’.
The results are shown in Table \ref{tab:compare_veto_mf}.
Our DRM consistently and significantly outperforms the recent VETO+MEET in terms of  both ng-M50/100 and ng-F@50/100 metrics. 
These results indicate the consistent effectiveness of our DRM under the setting of ``without graph constraint’’.

\begin{table*}[tbp]
    \centering
    \setcounter{table}{5}
    \scalebox{1.0}{
            \begin{tabular}{ l | c c  | c   c | c  c}
                \hline
                \multirow{2}{*}{Models} & \multicolumn{2}{c|}{PredCls} & \multicolumn{2}{c|}{SGCls} & \multicolumn{2}{c}{SGDet} \\
                & M@50/100 & F@50/100 
               & M@50/100 & F@50/100 
               & M@50/100 & F@50/100  \\
                \hline
                \hline
                IMP~\cite{xu2017scene}$_{\textit{CVPR'17}}$
                & 36.1 / 37.5 & 18.6 / 19.9
                & 21.9 / 22.5 & 10.6 / 11.1
                & 15.1 / 18.3 & \, 7.2 / 9.1 \,  \\
                VTransE~\cite{zhang2017visual}$_{\textit{CVPR'17}}$
                & 40.2 / 41.7 & 24.0 / 25.6
                & 23.4 / 24.1 & 13.5 / 14.3
                & 17.4 / 20.2 & \, 8.6 / 10.4   \\
                MOTIFS~\cite{zellers2018neural}$_{\textit{CVPR'18}}$
                & 40.3 / 41.9 & 23.9 / 25.6
                & 23.6 / 24.2 & 13.3 / 14.0
                & 18.8 / 21.9 & \, 9.4 / 11.5   \\
                G-RCNN~\cite{yang2018graph}$_{\textit{ECCV'18}}$
                & 40.9 / 42.2 & 26.2 / 27.4
                & 23.0 / 24.0 & 14.5 / 15.2
                & 17.8 / 19.7 & \, 9.7 / 11.0  \\
                VCTREE~\cite{tang2019learning}$_{\textit{CVPR'19}}$
                & 41.1 / 42.7 & 26.6 / 28.3
                & 24.1 / 25.0 & 13.2 / 13.8
                & 19.2 / 21.7 & 10.7 / 12.1  \\
                GPS-Net~\cite{lin2020gps}$_{\textit{CVPR'20}}$
                & 40.2 / 41.9 & 24.7 / 26.6
                & 23.2 / 24.2 & 13.9 / 14.8
                & 19.0 / 22.3 & 11.0 / 13.9  \\
                RU-Net~\cite{lin2022ru}$_{\textit{CVPR'22}}$
                & \, \; - \, / 46.9 & \, \; - \, / 35.9
                & \, \; - \, / 29.0 & \, \; - \, / 21.8
                & \, \; - \, / 24.2 & \, \; - \, / 16.8  \\
                HL-Net~\cite{lin2022hl}$_{\textit{CVPR'22}}$
                & \, \; - \, / 45.9 & \, \; - \, / 34.3
                & \, \; - \, / 28.5 & \, \; - \, / 20.6
                & \, \; - \, / 23.7 & \, \; - \, / 14.8  \\
                PE-Net(P)~\cite{zheng2023prototype}$_{\textit{CVPR'23}}$
                & 45.7 / 47.8 & 34.5 / 37.3
                & 27.2 / 28.6 & 19.9 / 21.9
                & 20.7 / 24.0 & 14.0 / 16.9  \\
                VETO~\cite{sudhakaran2023vision}$_{\textit{ICCV'23}}$
                & 43.5 / 45.5 & 33.6 / 36.0
                & 23.4 / 24.4 & 16.9 / 18.0
                & 17.8 / 20.5 & 12.5 / 14.6  \\
                \hline
                TDE$^{\diamond}$~\cite{tang2020unbiased}$_{\textit{CVPR'20}}$
                & 35.9 / 40.3 & 32.9 / 37.2
                & 20.4 / 22.4 & 17.8 / 19.9
                & 12.6 / 15.1 & 11.0 / 13.2  \\
                CogTree$^{\diamond}$~\cite{yu2021cogtree}$_{\textit{IJCAI'21}}$
                & 31.0 / 32.9 & 30.3 / 32.4
                & 18.3 / 19.2 & 17.6 / 18.7
                & 15.2 / 17.0 & 13.7 / 15.4  \\
                BPL-SA$^{\diamond}$~\cite{guo2021general}$_{\textit{ICCV'21}}$
                & 40.2 / 42.1 & 37.5 / 39.5
                & 23.3 / 24.3 & 21.3 / 22.4
                & 18.3 / 21.3 & 17.0 / 19.7  \\
                NICE$^{\diamond}$~\cite{li2022devil}$_{\textit{CVPR'22}}$
                & 42.5 / 44.8 & 38.8 / 41.3
                & 24.9 / 26.0 & 22.1 / 23.5
                & 20.0 / 23.1 & 17.0 / 19.8  \\
                PPDL$^{\diamond}$~\cite{li2022ppdl}$_{\textit{CVPR'22}}$
                & 39.7 / 40.5 & 38.3 / 39.2
                & 23.0 / 23.8 & 21.7 / 22.5
                & 16.3 / 18.7 & 14.8 / 17.3  \\
                GCL$^{\diamond}$~\cite{dong2022stacked}$_{\textit{CVPR'22}}$
                & 39.4 / 41.3 & 39.1 / 41.1
                & 23.5 / 24.5 & 23.2 / 24.2
                & 17.6 / 20.7 & 17.6 / 20.6  \\
                INF$^{\diamond}$~\cite{biswas2023probabilistic}$_{\textit{CVPR'23}}$
                & 38.1 / 42.9 & 33.4 / 39.4
                & 23.4 / 25.6 & 20.0 / 23.0
                & 16.7 / 19.4 & 13.5 / 16.3  \\
                CFA$^{\diamond}$~\cite{li2023compositional}$_{\textit{ICCV'23}}$
                & 44.9 / 47.4 
                & \textcolor{blue}{\textbf{\underline{43.0}}} / \textcolor{blue}{\textbf{\underline{45.6}}}
                & 26.0 / 27.3 & 22.9 / 24.4
                & 20.3 / 23.7 & 17.8 / 20.8  \\
                EICR$^{\diamond}$~\cite{min2023environment}$_{\textit{ICCV'23}}$
                & 45.1 / 47.2 & 42.8 / 45.0
                & 27.7 / 28.6 
                & \textcolor{blue}{\textbf{\underline{26.0}}} / \textcolor{blue}{\textbf{\underline{27.0}}}
                & \textcolor{red}{\textbf{21.7}} / \textcolor{red}{\textbf{25.2}} 
                & \textcolor{red}{\textbf{19.9}} / \textcolor{blue}{\textbf{\underline{23.3}}}  \\
                BGNN~\cite{li2021bipartite}$_{\textit{CVPR'21}}$
                & 44.8 / 47.1 & 40.2 / 42.8
                & 25.9 / 27.5 & 20.7 / 23.1
                & 20.9 / 24.2 & 15.9 / 18.6  \\
                SHA+GCL~\cite{dong2022stacked}$_{\textit{CVPR'22}}$
                & 38.4 / 40.7 & 38.1 / 40.4
                & 22.9 / 24.1 & 22.9 / 24.1
                & 16.4 / 19.6 & 16.3 / 19.5  \\
                PE-Net~\cite{zheng2023prototype}$_{\textit{CVPR'23}}$
                & \textcolor{red}{\textbf{48.2}} / \textcolor{red}{\textbf{50.5}} 
                & 42.4 / 45.0
                & \textcolor{blue}{\textbf{\underline{28.6}}} / \textcolor{blue}{\textbf{\underline{29.8}}}
                & 24.5 / 25.8
                & 21.6 / 24.9 & 17.7 / 20.5  \\
                SQUAT~\cite{jung2023devil}$_{\textit{ICCV'23}}$
                & 43.3 / 45.7 & 39.7 / 42.4
                & 25.3 / 26.6 & 22.9 / 24.3
                & 19.3 / 22.7 & 17.9 / 21.0  \\
                \hline
                \textbf{DRM} w/o \textbf{DKT} 
                & \textcolor{blue}{\textbf{\underline{46.8}}} / \textcolor{blue}{\textbf{\underline{48.9}}}
                & 35.0 / 37.8
                & \textcolor{red}{\textbf{28.9}} / \textcolor{red}{\textbf{29.9}} 
                & 20.7 / 22.1
                & \textcolor{blue}{\textbf{\underline{21.5}}} / \textcolor{blue}{\textbf{\underline{25.1}}}
                & 14.2 / 17.4  \\
                \textbf{DRM} 
                & 45.5 / 47.7 
                & \textcolor{red}{\textbf{45.4}} / \textcolor{red}{\textbf{47.6}}
                & 27.7 / 28.8 
                & \textcolor{red}{\textbf{27.6}} / \textcolor{red}{\textbf{28.8}}
                & 19.7 / 23.5 
                & \textcolor{blue}{\textbf{\underline{19.7}}} / \textcolor{red}{\textbf{23.5}}  \\
                \hline
            \end{tabular}
    }
    \caption{
    Results in terms of M@K and F@K for three tasks on the VG150 dataset with graph constraints.
    ``$\diamond$'' denotes the combination of MOTIFS with a model-agnostic unbiasing strategy. The best and second best results under each setting are respectively marked in \textcolor{red}{\textbf{red}} and \textcolor{blue}{\textbf{\underline{underline blue}}}.
    }
    \label{tab:compare_vg_mf}
\end{table*}

\begin{table*}[tbp]
    \centering
     \scalebox{1.0}{
    % \resizebox{\textwidth}{!}{
        % \setlength{\tabcolsep}{2mm}{
        % \resizebox{14cm}{!}{
            \begin{tabular}{ l | c  c | c   c | c c}
                \hline
                \multirow{2}{*}{Models} & \multicolumn{2}{c|}{PredCls} & \multicolumn{2}{c|}{SGCls} & \multicolumn{2}{c}{SGDet} \\
                & M@50/100 & F@50/100 
               & M@50/100 & F@50/100
               & M@50/100 & F@50/100\\
                \hline
                \hline
                VTransE~\cite{zhang2017visual}$_{\textit{CVPR'17}}$
                & 34.9 / 36.5 & 22.4 / 23.8
                & 20.8 / 21.5 & 13.0 / 13.9
                & 16.5 / 18.7 & 9.6 / 10.9  \\
                MOTIFS~\cite{zellers2018neural}$_{\textit{CVPR'18}}$
                & 40.9 / 42.0 & 26.2 / 27.2
                & 21.2 / 21.8 & 13.2 / 13.8
                & 17.7 / 20.4 & 10.5 / 12.5  \\
                VCTREE~\cite{tang2019learning}$_{\textit{CVPR'19}}$
                & 40.2 / 41.6 & 26.3 / 27.5
                & 21.0 / 21.6 & 12.8 / 13.4
                & 17.4 / 19.7 & 10.6 / 12.0  \\
                SHA~\cite{dong2022stacked}$_{\textit{CVPR'22}}$
                & 41.4 / 43.2 & 29.8 / 31.9
                & 20.6 / 21.3 & 13.5 / 14.2
                & 16.1 / 18.5 & 10.5 / 12.3  \\
                VETO~\cite{sudhakaran2023vision}$_{\textit{ICCV'23}}$
                & \textcolor{red}{\textbf{42.9}} / \textcolor{red}{\textbf{44.1}}
                & 31.9 / 33.1
                & 19.5 / 20.3 & 13.4 / 14.1
                & 16.6 / 18.6 & 11.0 / 12.7  \\
                \hline
                VTransE+GCL~\cite{dong2022stacked}$_{\textit{CVPR'22}}$
                & 33.0 / 34.9 & 32.8 / 34.7
                & 19.8 / 20.5 & 19.2 / 20.0
                & 15.0 / 17.2 & 15.0 / 17.2  \\
                MOTIFS+GCL~\cite{dong2022stacked}$_{\textit{CVPR'22}}$
                & 40.6 / 42.2 & 40.2 / 41.8
                & 20.3 / 21.1 & 19.8 / 20.6
                & 17.7 / 20.3 
                & \textcolor{blue}{\textbf{\underline{17.6}}} / \textcolor{blue}{\textbf{\underline{20.2}}}  \\
                VCTREE+GCL~\cite{dong2022stacked}$_{\textit{CVPR'22}}$
                & 40.1 / 41.7 & 39.5 / 41.1
                & 20.5 / 21.3 & 20.0 / 20.8
                & 16.6 / 19.3 & 16.5 / 19.1  \\
                SHA+GCL~\cite{dong2022stacked}$_{\textit{CVPR'22}}$
                & 41.9 / 43.6 
                & \textcolor{blue}{\textbf{\underline{41.8}}} / \textcolor{blue}{\textbf{\underline{43.6}}}
                & 21.0 / 21.8 
                & \textcolor{blue}{\textbf{\underline{21.0}}} / \textcolor{blue}{\textbf{\underline{21.7}}}
                & 16.3 / 19.0 & 16.2 / 18.9  \\
                \hline
                \textbf{DRM} w/o \textbf{DKT} 
                & 42.5 / 43.7 & 28.5 / 29.7
                & \textcolor{red}{\textbf{21.8}} / \textcolor{red}{\textbf{22.3}} 
                & 11.9 / 12.3
                & \textcolor{red}{\textbf{18.8}} / \textcolor{red}{\textbf{21.5}} & 11.3 / 13.5  \\
                \textbf{DRM}
                & \textcolor{blue}{\textbf{\underline{42.6}}} / \textcolor{blue}{\textbf{\underline{44.0}}}
                & \textcolor{red}{\textbf{42.5}} / \textcolor{red}{\textbf{43.9}}
                & \textcolor{blue}{\textbf{\underline{21.6}}} / \textcolor{blue}{\textbf{\underline{22.3}}}
                & \textcolor{red}{\textbf{21.5}} / \textcolor{red}{\textbf{22.2}}
                & \textcolor{blue}{\textbf{\underline{18.8}}} / \textcolor{blue}{\textbf{\underline{21.4}}}
                & \textcolor{red}{\textbf{18.7}} / \textcolor{red}{\textbf{21.3}}  \\
                \hline
            \end{tabular}
    }
    \caption{
    Results in terms of M@K and F@K for three tasks on the GQA200 dataset with graph constraints.
   The best and second best results under each setting are respectively marked in \textcolor{red}{\textbf{red}} and \textcolor{blue}{\textbf{\underline{underline blue}}}.
    }
    \label{tab:compare_gqa_mf}
\end{table*}

\begin{table*}[tbp]
    \centering 
     \scalebox{0.95}{
            \begin{tabular}{ c | l | c c | c c | c c}
                \hline
                 \multirow{3}{*}{Datasets} & \multirow{3}{*}{Models} &  \multicolumn{2}{c|}{PredCls} 
                & \multicolumn{2}{c|}{SGCls} & \multicolumn{2}{c}{SGDet} \\
                & ~ & ng- & ng- & ng- 
                & ng- & ng- & ng- \\
               & ~ & R@50/100 & mR@50/100 
               & R@50/100 & mR@50/100 
               & R@50/100 & mR@50/100 \\
                \hline
                \hline
                \multirow{3}{*}{VG150}
                & VETO+MEET~\cite{sudhakaran2023vision}$_{\textit{ICCV'23}}$
                & 74.0 / 78.9 & 42.0 / 52.4 
                & 41.1 / 44.0 & 22.3 / 27.4  
                & 28.6 / 34.0 & 10.6 / 13.8  \\
                ~ & \textbf{DRM} w/o \textbf{DKT}  
                & \textbf{85.8} / \textbf{92.0} & 42.8 / 57.1 
                & \textbf{53.6} / \textbf{56.8} & 24.3 / 32.1
                & \textbf{37.4} / \textbf{43.9} & 13.8 / 19.1 \\
                ~ & \textbf{DRM} 
                & 68.1 / 80.4 & \textbf{62.9} / \textbf{73.5} 
                & 43.2 / 50.1 & \textbf{37.6} / \textbf{43.6}
                & 26.1 / 33.6 & \textbf{25.0} / \textbf{30.5} \\
                \hline
                \hline
                \multirow{3}{*}{GQA200}
                & VETO+MEET~\cite{sudhakaran2023vision}$_{\textit{ICCV'23}}$
                & 73.9 / 78.3 & 43.3 / 50.5 
                & 34.6 / 37.2 & 19.7 / 22.5
                & 26.7 / 31.0 & 12.1 / 16.0 \\
                ~ & \textbf{DRM} w/o \textbf{DKT} 
                & \textbf{79.6} / \textbf{85.9} & 44.8 / 59.4
                & \textbf{43.1} / \textbf{46.4} & 17.9 / 24.1 
                & \textbf{33.1} / \textbf{38.4} & 13.2 / 18.6 \\
                ~ & \textbf{DRM}
                & 68.5 / 78.2 & \textbf{65.6} / \textbf{75.7}
                & 36.6 / 41.8 & \textbf{32.1} / \textbf{36.9}
                & 24.4 / 30.4 & \textbf{26.0} / \textbf{30.8} \\
                \hline
            \end{tabular}
    }
    \caption{
    Comparison results with VETO+MEET on the VG150 and GQA200 datasets without graph constraint. 
    The metrics ``ng-R@K’’ and ``ng-mR@K’’ denote the No Graph Constraint Recall@K and No Graph Constraint Mean Recall@K, respectively.
   The best results under each setting are respectively marked in \textbf{bold}.
    }
    \label{tab:compare_veto_un}
\end{table*}

\begin{table*}[tbp]
    \centering
     \scalebox{1.0}{
            \begin{tabular}{ c | l | c c | c c | c c}
                \hline
                 \multirow{3}{*}{Datasets} & \multirow{3}{*}{Models} &  \multicolumn{2}{c|}{PredCls} 
                & \multicolumn{2}{c|}{SGCls} & \multicolumn{2}{c}{SGDet} \\
                ~& ~ & ng- & ng- & ng- 
                & ng- & ng- & ng- \\
               ~ & ~ & M@50/100 & F@50/100 
               & M@50/100 & F@50/100 
               & M@50/100 & F@50/100 \\
                \hline
                \hline
                \multirow{3}{*}{VG150}
                & VETO+MEET~\cite{sudhakaran2023vision}$_{\textit{ICCV'23}}$
                & 58.0 / 65.7 & 53.6 / 63.0
                & 31.7 / 35.7 & 28.9 / 33.8
                & 19.6 / 23.9 & 15.5 / 19.6 \\
                ~ & \textbf{DRM} w/o \textbf{DKT}  
                & 64.3 / 74.6 & 57.1 / 70.5
                & 39.0 / 44.5 & 33.4 / 41.0
                & 25.6 / 31.5 & 20.2 / 26.6 \\
                ~ & \textbf{DRM} 
                & \textbf{65.5} / \textbf{77.0} & \textbf{65.4} / \textbf{76.8}
                & \textbf{40.4} / \textbf{46.9} & \textbf{40.2} / \textbf{46.6}
                & \textbf{25.6} / \textbf{32.1} & \textbf{25.5} / \textbf{32.0} \\
                \hline
                \hline
                \multirow{3}{*}{GQA200}
                & VETO+MEET~\cite{sudhakaran2023vision}$_{\textit{ICCV'23}}$
                & 58.6 / 64.4 & 54.6 / 61.4
                & 27.2 / 29.9 & 25.1 / 28.0
                & 19.4 / 23.5 & 16.7 / 21.1 \\
                ~ & \textbf{DRM} w/o \textbf{DKT} 
                & 62.2 / 72.7 & 57.3 / 70.2
                & 30.5 / 35.3 & 25.3 / 31.7
                & 23.2 / 28.5 & 18.9 / 25.1 \\
                ~ & \textbf{DRM}
                & \textbf{67.1} / \textbf{77.0} & \textbf{67.0} / \textbf{76.9}
                & \textbf{34.4} / \textbf{39.4} & \textbf{34.2} / \textbf{39.2}
                & \textbf{25.2} / \textbf{30.6} & \textbf{25.2} / \textbf{30.6} \\
                \hline
            \end{tabular}
    }
    \caption{
    Results in terms of ng-M@K and ng-F@K for three tasks on the VG150 and GQA200 datasets without graph constraints. The metrics
    ``ng-M@K'' and ``ng-F@K'' denote the No Graph Constraint Mean@K and No Graph Constraint Harmonic Mean@K, respectively.
   The best results under each setting are respectively marked in \textbf{bold}.
    }
    \label{tab:compare_veto_mf}
\end{table*}

\begin{figure*}[!tb]
\setcounter{figure}{4}
\centering {\includegraphics[width=1.0\textwidth]{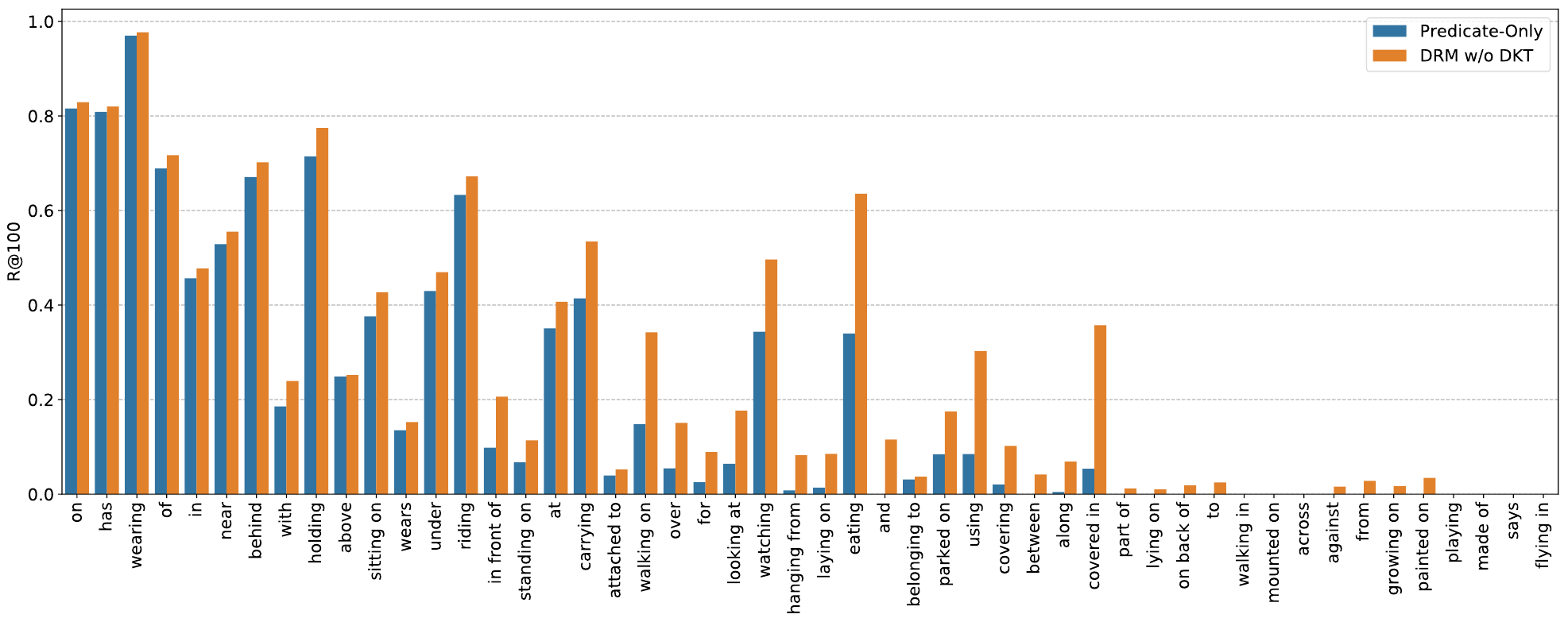}}
\caption{
Results in terms of Recall@100 of all predicate classes of Predicate-Only and DRM w/o DKT on the PredCls task. Predicates are sorted according to their frequency.
}
\label{fig:r100_prd}
\end{figure*}

% \begin{figure*}[!tb]
% \centering {\includegraphics[width=1.0\textwidth]{mR_bar.pdf}}
% \caption{
% Recall@100 of all predicate classes of DRM w/o DKT and DRM on the PredCls task. Predicates are sorted according to their frequency.
% }
% \label{fig:framework}
% \end{figure*}

\begin{figure*}[!tb]
\centering {\includegraphics[width=1.0\textwidth]{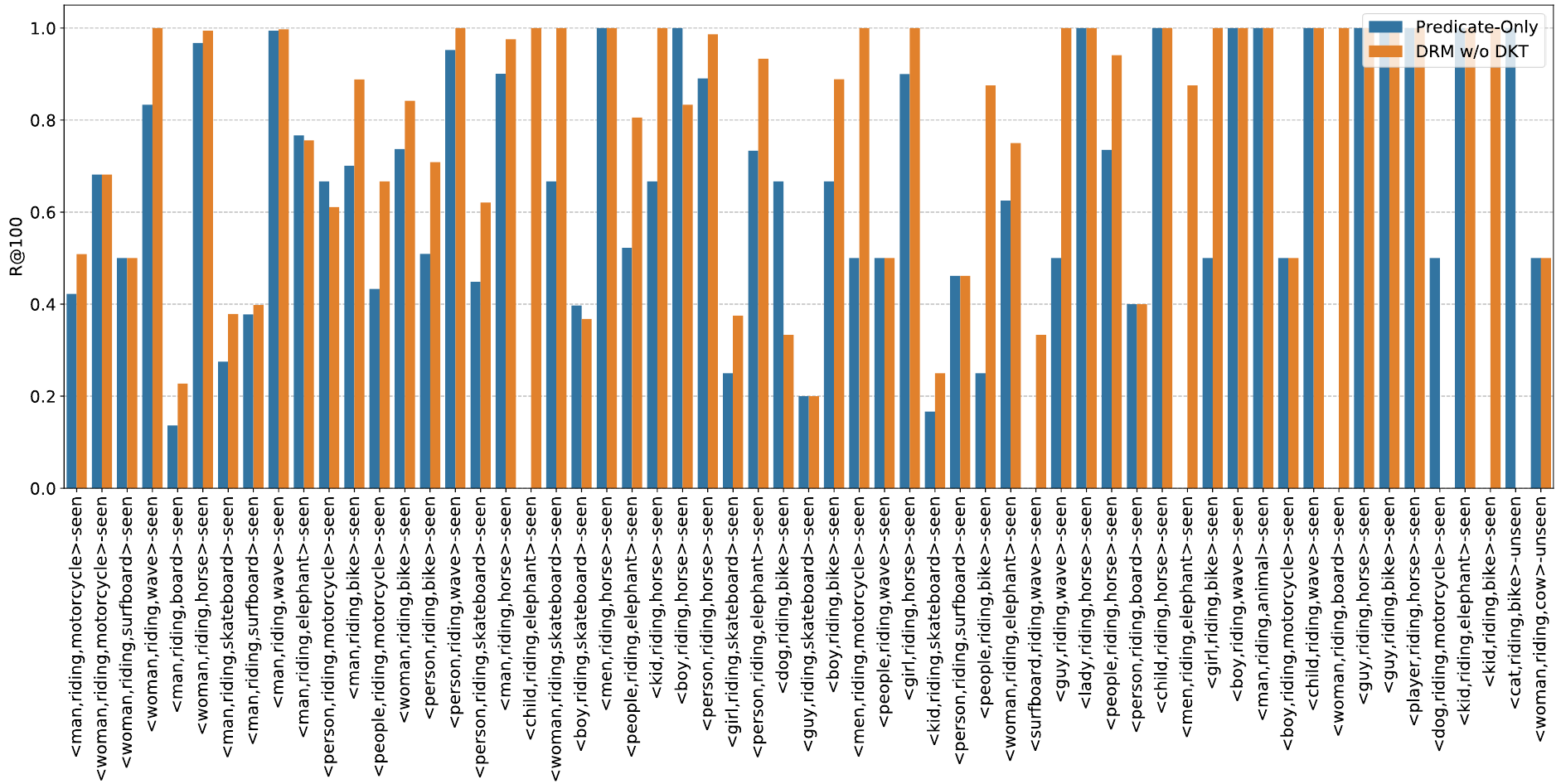}}
\caption{
Results in terms of Recall@100 for triplets belonging to predicate ``riding'' of Predicate-Only and DRM w/o DKT on the PredCls task. 
The terms ``seen'' and ``unseen'' represent whether the triplets appear in the training set or not, respectively.
}
\label{fig:r100_tpt_ride}
\end{figure*}

\begin{figure*}[!tb]
\centering {\includegraphics[width=1.0\textwidth]{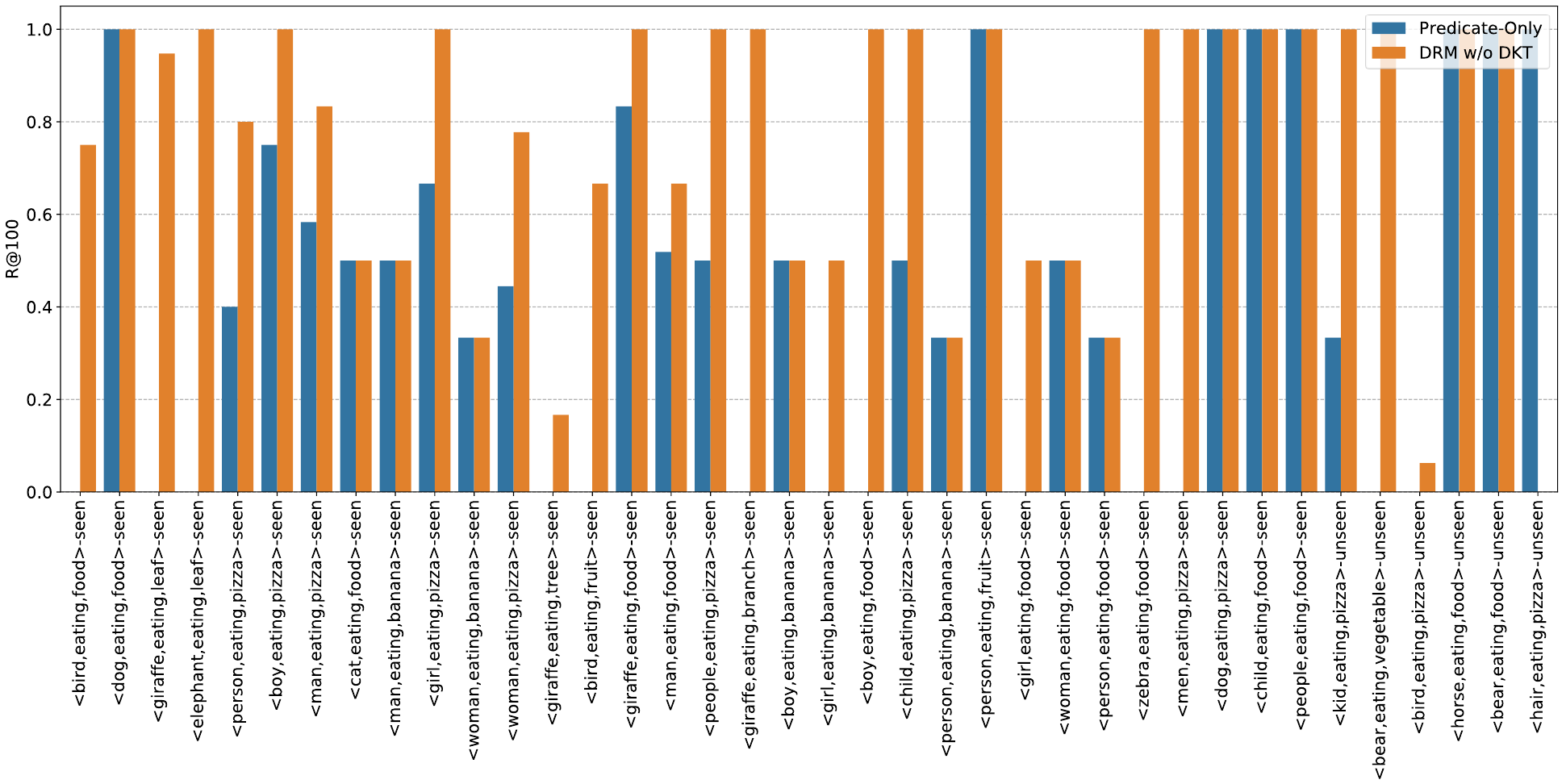}}
\caption{
Results in terms of Recall@100 for triplets belonging to predicate ``eating'' of Predicate-Only and DRM w/o DKT on the PredCls task. 
The terms ``seen'' and ``unseen'' represent whether the triplets appear in the training set or not, respectively. 
}
\label{fig:r100_tpt_eat}
\end{figure*}

\begin{table*}[tbp]
    \centering
    \scalebox{1.0}{
            \begin{tabular}{ l | c c  | c   c | c  c}
                \hline
                \multirow{2}{*}{Models} & \multicolumn{2}{c|}{PredCls} & \multicolumn{2}{c|}{SGCls} & \multicolumn{2}{c}{SGDet} \\
                & mR@50 & mR@100 
               & mR@50 & mR@100 
               & mR@50 & mR@5100  \\
                \hline
                \hline
                PE-Net~\cite{zheng2023prototype}$_{\textit{CVPR'23}}$
                & 31.4 &  33.5 
                &  18.2 & 19.3
                & 12.3 &  14.3  \\
                \textbf{PE-Net} + \textbf{DKT-P}
                & \textbf{44.3} & \textbf{48.4} 
                & \textbf{24.9}  & \textbf{27.0} 
                & \textbf{15.0} & \textbf{17.7}  \\
                \hline
            \end{tabular}
    }
    \caption{
    The results on three tasks of PE-Net equipped with our DKT on VG150 dataset.
    }
    \label{tab:ablation_penet}
\end{table*}

\begin{table}[tbp]
    \centering
         \scalebox{0.95}{
            \begin{tabular}{ c c | c c | c c }
                \hline
                \multirow{2}{*}{$\tau_p$} & 
                \multirow{2}{*}{$\tau_t$} & 
                \multicolumn{4}{c}{PredCls}  \\
                \cline{3-6}
                ~ & ~
                & R@50 & R@100 & mR@50 & mR@100 \\
                \hline
                \hline
                 0.1 & 0.1
                & 70.1  & 71.9 & 22.6 & 24.9 \\
                 0.2 & 0.1
                & \textbf{70.2}  & \textbf{72.1} & \textbf{23.3} & \textbf{25.6} \\
                 0.3 & 0.1
                & 70.1  & 72.0 & 22.5 & 24.9 \\
                 0.2 & 0.2
                & 70.2  & 72.0 & 23.0 & 25.3 \\
                \hline
            \end{tabular}
        }
    \caption{
    Hyper-parameters analysis of the temperature $\tau_p$ and $\tau_t$.
    }
    \label{tab:ablation_tau}
\end{table}

\begin{table}[tbp]
    \centering
         \scalebox{0.95}{
            \begin{tabular}{ c c | c c | c c }
                \hline
                \multirow{2}{*}{$\lambda_p$} & 
                \multirow{2}{*}{$\lambda_e$} & 
                \multicolumn{4}{c}{SGCls}  \\
                \cline{3-6}
                ~ & ~
                & R@50 & R@100 & mR@50 & mR@100 \\
                \hline
                \hline
                3 & 0.1
                & 43.4  & 44.3 & 13.1 & 14.3 \\
                3 & 0.2
                & 43.9  & 44.8 & 13.3 & 14.5 \\
                3 & 0.5
                & \textbf{44.3}  & \textbf{45.2} & \textbf{13.5} & \textbf{14.6} \\
                3 & 0.8
                & 44.2  & 45.1 & 13.0 & 14.2 \\
                \hline
                2 & 0.5
                & 44.2  & 45.2 & 12.9 & 14.1 \\
                3 & 0.5
                & \textbf{44.3}  & \textbf{45.2} & \textbf{13.5} & \textbf{14.6} \\
                4 & 0.5
                & 44.2  & 45.1 & 13.4 & 14.6 \\
                \hline
            \end{tabular}
        }
    \caption{
    Hyper-parameters analysis of the loss weights $\lambda_p$ and $\lambda_t$.
    }
    \label{tab:ablation_lambda}
\end{table}

\begin{figure*}[!tb]
\centering {\includegraphics[width=1.0\textwidth]{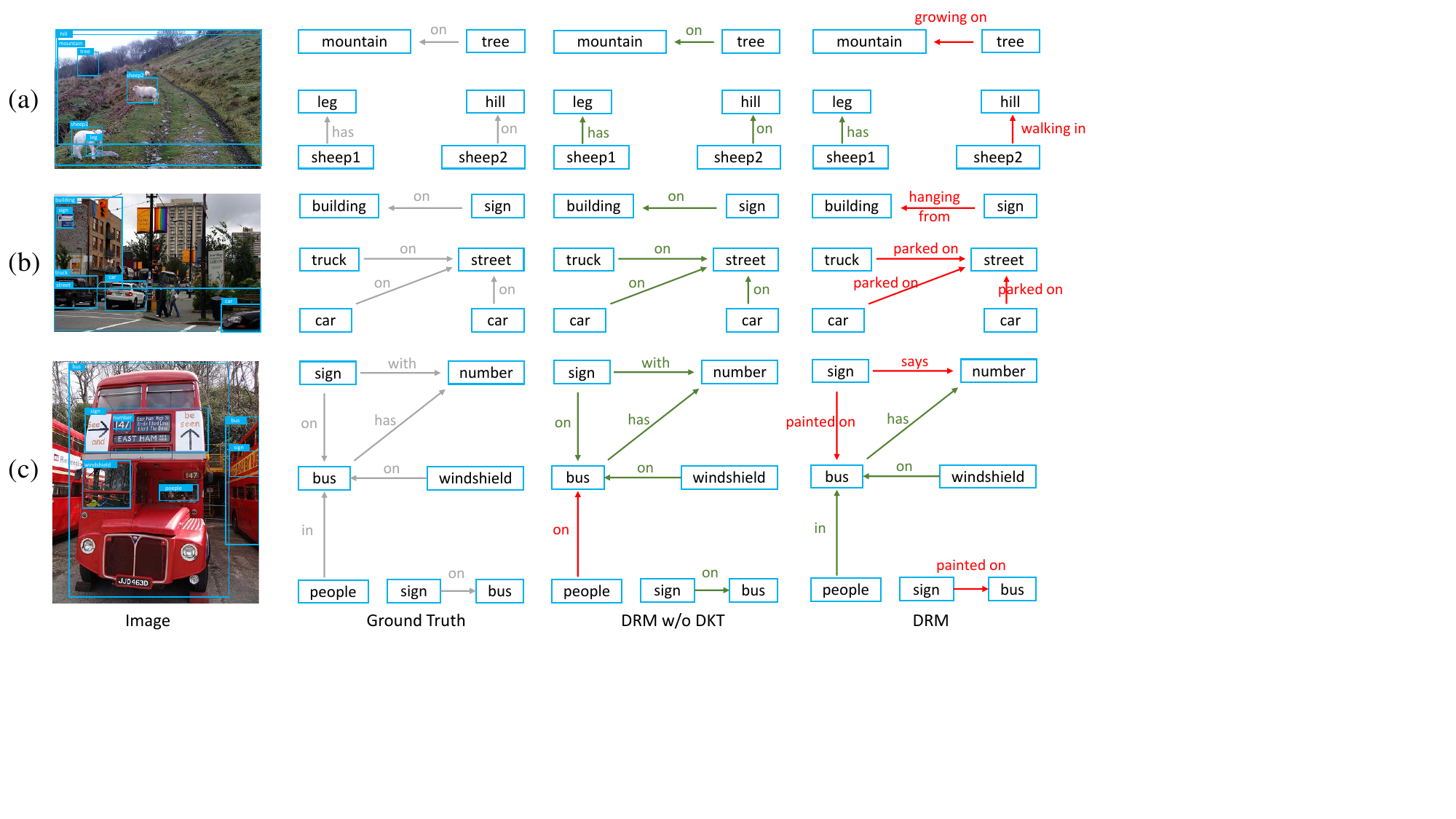}}
\caption{
Scene graphs generated by our DRM w/o DKT and DRM in the PredCls Task. DRM tends to generate more precise fine-grained relations than DRM w/o DKT, which leads to the performance degradation on R@K.
The use of green and red colors indicates whether the prediction matches the ground truth or not, respectively.
}
\label{fig:vis_1}
\end{figure*}

\begin{figure*}[!tb]
\centering {\includegraphics[width=1.0\textwidth]{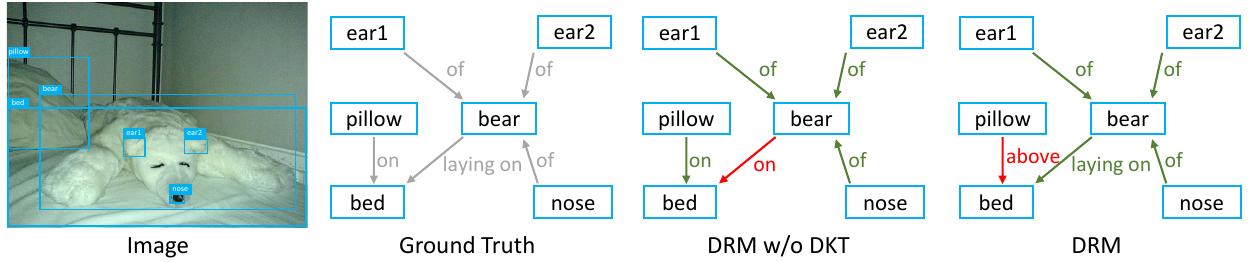}}
\caption{
Scene graphs generated by our DRM w/o DKT and DRM in the PredCls Task. Triplet \textit{\textless bear, laying on, bed\textgreater } only appears once in the VG150 training set.
The use of green and red colors indicates whether the prediction matches the ground truth or not, respectively.
}
\label{fig:vis_2}
\end{figure*}

\section{Additional Ablation Studies}
We conduct additional ablation studies to further evaluate the effectiveness of our dual-granularity learning and DKT strategy.
We combine DKT with a recent state-of-the-art model, PE-NET \cite{zheng2023prototype}. 
As PE-Net is only concerned with modeling predicate features, we only transfer its predicate knowledge from head to tail. 
As shown in Table \ref{tab:ablation_penet}, the incorporation of DKT substantially boosts the performance of PE-NET in three tasks at mR@K.

To further demonstrate the effectiveness of our method in modeling triplet clues, we provide the R@100 performance of our methods ``predicate-only'' and ``DRM w/o DKT'' on each predicate. 
As shown in Figure \ref{fig:r100_prd}, our method outperforms the ``predicate-only'' baseline on each predicate.
We also present the performance of the Recall@100 for the predicate ``riding'' and the predicate ``eating'' at the fine-grained triplet level. 
The results are illustrated in Figures \ref{fig:r100_tpt_ride} and \ref{fig:r100_tpt_eat}.
We can observe that the performance is significantly improved in the seen triplets, especially for 
\textit{\textless giraffe, eating, leaf\textgreater } and \textit{\textless bird, eating, fruit\textgreater }, 
where our method can accurately predict multiple eating expressions that cannot be captured using only predicates. 
This suggests that our method is able to learn the triplet cues in the training, and utilize them to reason about the relationships under specific subject-object pairs during inference.

\section{Hyper-parameters analysis}
We first investigate the impact of hyper-parameters $\tau_p$ and $\tau_t$ in the Dual-granularity Constraints.
The results are illustrated in Table \ref{tab:ablation_tau}.
The decrease in $\tau_p$ and $\tau_t$ leads to the more compact predicate and triplet representations. 
We observe that the model achieves the best performance when $\tau_p=0.2$ and $\tau_t=0.1$, indicating a compact aggregation of the triplet compared to the predicate. 
This observation aligns with the fact that variations in triplet are significantly smaller than those in the predicate.

During pre-training, the predicate classification loss is much smaller than the entity classification loss due to the long-tailed distribution of predicates. 
To balance the scales between these losses, we set $\lambda_p$ to be larger than $\lambda_e$. 
Experimental results in Table \ref{tab:ablation_lambda} show that the model performs best when $\lambda_p=3$ and $\lambda_e=0.5$.

\section{Visualization Results}
We present a visualization of the results of our DRM w/o DKT and our DRM on the PredCls task on the VG150 dataset. 
The results are shown in Figure \ref{fig:vis_1} and \ref{fig:vis_2}.
As the VG dataset is incompletely labeled, we focus our analysis only on labeled relations in the dataset.
We observe that DRM w/o DKT can accurately predict the ground-truth relationships. However, our DRM often predicts these relationships differently. 
This difference arises from the tendency of our DRM to predict coarse-grained relations as more precise fine-grained relations. 
For instance, for the subject-object pair ``tree-mountain'', DRM generates the more accurate predicate ``growing on'' while the DRM w/o DKT predicts the coarse-grained predicate ``on''.
These fine-grained predictions lead to a decrease in R@K and an increase in mR@K. 
This degradation in R@K is an inevitable result due to the large number of coarsely labeled relations in the test dataset.

\end{document}